\documentclass[11pt,a4paper]{article}
\usepackage[hyperref]{naaclhlt2019}
\usepackage{natbib}
\usepackage{booktabs}
\usepackage{times}
\usepackage{latexsym}
\usepackage{amsmath}
\usepackage{amsfonts}
\usepackage{mathtools}
\usepackage{url}
\usepackage{graphicx}
\usepackage{xcolor}
\usepackage{appendix}
\usepackage{rotating}
\usepackage{multirow}

\usepackage{color, colortbl}
\definecolor{Gray}{gray}{0.9}
\usepackage[first=0,last=9]{lcg}

\aclfinalcopy 

\newcommand{\smalltt}[1]{{{\small \texttt{#1}}}}

\title{
What makes a good conversation? \\
How controllable attributes affect human judgments
}

\author{
  Abigail See\thanks{\enspace A.S. completed most of this work at Facebook (FAIR).} \\
  Stanford University \\
  {\tt \small{abisee@stanford.edu}} \\ \And
  Stephen Roller \\
  Facebook AI Research \\
  {\tt \small{roller@fb.com}} \\\And
  Douwe Kiela \\
  Facebook AI Research \\
  {\tt \small{dkiela@fb.com}} \\\And
  Jason Weston \\
  Facebook AI Research \\
  {\tt \small{jase@fb.com}} \\
}

\date{}

\begin{document}
\maketitle
\begin{abstract}
A good conversation requires balance -- between simplicity and detail; staying on topic and changing it; asking questions and answering them. 
Although dialogue agents are commonly evaluated via human judgments of overall quality, the relationship between quality and these individual factors is less well-studied. In this work, we examine two controllable neural text generation methods, conditional training and weighted decoding, in order to control four important attributes for chitchat dialogue: repetition, specificity, response-relatedness and question-asking. 
We conduct a large-scale human evaluation to measure the effect of these control parameters on multi-turn interactive conversations on the PersonaChat task. 
We provide a detailed ~~~analysis~of their relationship to high-level aspects of conversation, and show that by controlling combinations of these variables our models obtain clear improvements in human quality judgments.
\end{abstract}

\section{Introduction}
\label{sec:intro}
Neural generation models for dialogue, despite their ubiquity in current research, are still poorly understood. Well known problems, such as the genericness and repetitiveness of responses \citep{serban2016generative}, remain without a de facto solution. Strikingly, the factors that determine human judgments of overall conversation quality are almost entirely unexplored. Most works have been limited to the next utterance prediction problem, whereas a multi-turn evaluation is necessary to evaluate the quality of a full conversation.

\begin{figure}[t]
    \centering
    \includegraphics[width=\columnwidth]{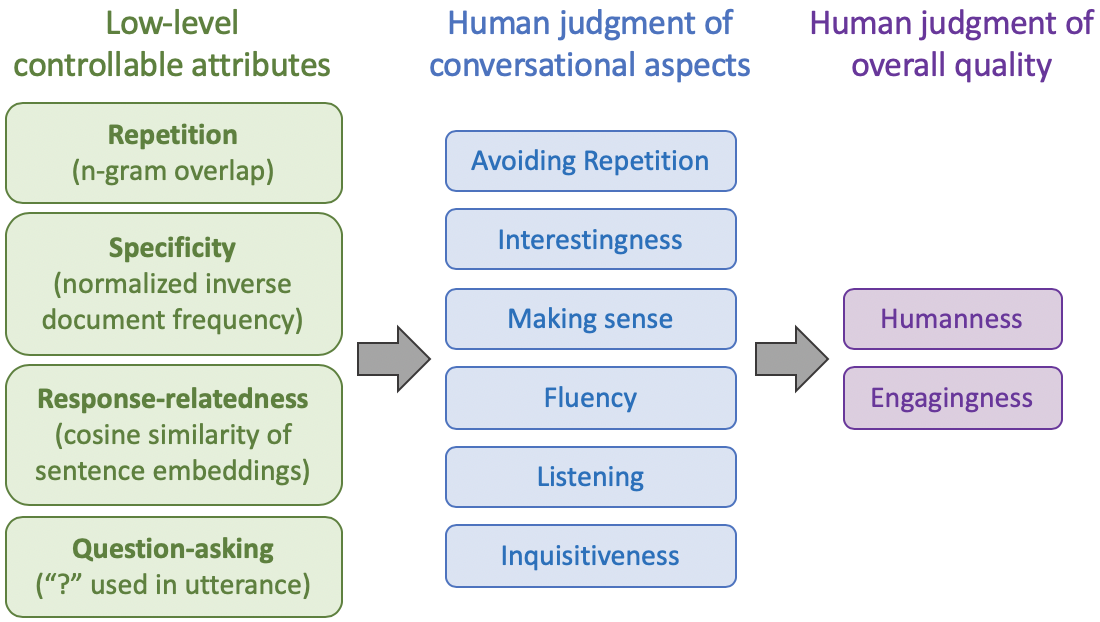}
    \caption{
        We manipulate four low-level attributes and measure their effect on human judgments of individual conversational aspects, as well as overall quality. 
    \label{fig:explain-the-paper}
    }
\end{figure}

In this work we both (i) conduct a large-scale study to identify the fine-grained factors governing human judgments of full conversations, and (ii) develop models that apply our findings in practice, leading to state-of-the-art performance. Specifically, we identify and study eight aspects of conversation that can be measured by human judgments, while varying four types of low-level attributes that can be algorithmically controlled in neural models; see Figure \ref{fig:explain-the-paper}. To control the low-level model attributes, we consider two simple but general algorithms: conditional training, in which the neural model is conditioned on additional control features, and weighted decoding, in which control features are added to the decoding scoring function at test time only. 

One major result of our findings is that existing work has ignored the importance of conversational flow, as standard models (i) repeat or contradict previous statements, (ii) fail to balance specificity with genericness, and (iii) fail to balance asking questions with other dialogue acts. Conducting experiments on the PersonaChat task \citep{Zhang2018PersonalizingToo}, we obtain significantly higher engagingness scores than the baseline by optimizing control of repetition, specificity and question-asking over multiple turns. Using these findings, our best model matches the performance of the winning entry in the recent NeurIPS ConvAI2 competition \citep{dinan2019second}, which was trained on much more data but had no control (see Section \ref{subsec:main_findings}). Our code, pretrained models, and full chatlogs, are available at \url{https://parl.ai/projects/controllable_dialogue}.
  
\section{Related Work}  
\label{sec:rel_work}
\paragraph{Dialogue}
\label{subsec:rel_work_dialog}
Dialogue evaluation is relatively well understood in goal-oriented tasks,
where automated approaches can be coded by measuring task completion 
\citep{bordes2016learning,asri2017frames,lemon,henderson2014second,wen2016network}.
Task success combined with dialogue cost can be linked to human judgments
like user satisfaction via the PARADISE framework \citep{walker1997paradise}.

However in chitchat tasks, which we study in this work,
automatic metrics and their relation to human ratings are less well-understood.
While word-overlap metrics are effective for question-answering and machine translation, for dialogue they have little to no correlation with human judgments \citep{liu2016not, novikova2017we} -- this is due to the open-ended nature of dialogue.
There are more recent attempts to find better automatic approaches,
such as adversarial evaluation \citep{li2017adversarial} and learning a scoring model 
\citep{lowe2017towards}, but their value is still unclear.

Nevertheless, a number of studies only use automatic metrics, 
with no human study at all \citep{lowe2015ubuntu,parthasarathi2018extending,serban2016building}.
Other works do use human evaluations \citep{dinan2018wizard,li2015diversity,li2016deep,venkatesh2018evaluating,vinyals2015neural,Zhang2018PersonalizingToo},
typically reporting just one type of judgment 
(either quality or appropriateness) 
via a Likert scale or pairwise comparison. 
Most of those works only consider single turn evaluations, often with a shortened dialogue history, rather than full multi-turn dialogue. 

A more comprehensive evaluation strategy has been studied within the scope of the 
Alexa prize \citep{venkatesh2018evaluating,guo2018topic} by combining
multiple automatic metrics designed to capture various conversational aspects (engagement, coherence, domain coverage, conversational depth and topical diversity).
Though these aspects have some similarity to the aspects studied here, we also focus on lower-level aspects (e.g. avoiding repetition, fluency), to understand how they correspond to both our controllable attributes, and to overall quality judgments.

 

\paragraph{Controllable neural text generation}
\label{subsec:rel_work_control}
Researchers have proposed several approaches to control aspects of RNN-based natural language generation such as sentiment, length, speaker style and tense 
\citep{fan2018controllable, ficler2017controlling, ghazvininejad2017hafez, hu2017toward, kikuchi2016controlling, peng2018towards, wang2017steering}.
In particular, several works use control to tackle the same common sequence-to-sequence problems we address here (particularly genericness and unrelated output), in the context of single-turn response generation \citep{baheti2018generating, li2015diversity, li2017learning, shen2017conditional, xing2017topic, zhang2018learning, zhou2017mechanism}.
By contrast, we focus on developing controls for, and human evaluation of, \textit{multi}-turn interactive dialogue -- this includes a new method (described in Section \ref{sec:control_methods}) to control attributes at the \textit{dialogue} level rather than the utterance level.


In this work, we require a control method that is both general-purpose (one technique to simultaneously control many attributes) and easily tunable (the control setting is adjustable after training).
Given these constraints, we study two control methods: conditional training (variants of which have been described by \cite{fan2018controllable, kikuchi2016controlling, peng2018towards}) and weighted decoding (described by \cite{ghazvininejad2017hafez} as a general technique, and by \cite{baheti2018generating} to control response-relatedness).
To our knowledge, this work is the first to systematically compare the effectiveness of two general-purpose control methods across several attributes.

\section{The PersonaChat dataset}
\label{sec:personachat}
PersonaChat \citep{Zhang2018PersonalizingToo} is a chitchat dialogue task involving two participants (two humans or a human and a bot).
Each participant is given a \textit{persona} -- a short collection of personal traits such as \textit{I'm left handed} or \textit{My favorite season is spring} -- and are instructed to get to know each other by chatting naturally using their designated personas, for 6--8 turns.
The training set contains 8939 conversations and 955 personas, collected via crowdworkers, plus 1000 conversations and 100 personas for validation, and a similar number in the hidden test set.
The PersonaChat task was the subject of the NeurIPS 2018 ConvAI2 Challenge \citep{dinan2019second}, in which competitors were first evaluated with respect to automatic metrics (perplexity, hits@1 and F1 score), and then with respect to human judgment via the question {\em ``How much did you enjoy talking to this user?"} on a scale of 1--4.

\section{Baseline model}
\label{sec:baseline}
Our baseline model is a 2-layer LSTM sequence-to-sequence model with attention. On any dialogue turn, the input $x$ to the encoder is the entire dialogue history (separated using unique speaker-identifying tokens), with the model's own persona prepended.
Conditioned on this input sequence $x$, the decoder generates a response $y$. Except when stated otherwise, all our models decode using beam search with beam size 20. 

We initialized the word embedding matrix with 300-dimensional GloVe embeddings \citep{pennington2014glove}. 
Using the ParlAI framework  \citep{miller2017parlai}, we pretrained the model on a dataset of 2.5 million Twitter message-response pairs,\footnote{The Twitter dataset is provided in ParlAI; details can be found here:  \url{https://parl.ai/docs/tasks.html}} then fine-tuned it on PersonaChat.
On the PersonaChat validation set, the baseline model has a perplexity of 26.83 and F1 of 17.02,
which would have placed us 4th out of 26 models in the ConvAI2 competition \citep{dinan2019second}.
We attempt to improve over this baseline using control.

\section{Controllable text generation methods}
\label{sec:control_methods}
Suppose we have a sequence-to-sequence model which gives $P(y|x) = \Pi_t P(y_t | x, y_1, \dots, y_{t-1})$, the conditional probability of a response $y$ (the model's next utterance) given input $x$ (the context, which in our case includes the model's own persona and the dialogue history).

Contrary to most previous work, which controls \textit{at the sentence level}, 
we wish to control attributes of the output $y$ \textit{at the dialogue level} -- meaning that a single control setting is used for a whole dialogue.
For example, to control question-asking, we provide a control setting at the beginning of each dialogue (e.g. \textit{20\% questions} or \textit{70\% questions}) rather than providing a control setting for each utterance (e.g. \textit{is a question} or \textit{isn't a question}).
With this approach, the sequence-to-sequence model is able to choose what value the controlled attribute should take for any particular utterance, but we are able to choose the overall distribution.
We find that this approach works well -- for example, the sequence-to-sequence model is generally good at detecting when to ask a question. In particular, this is easier than the alternative: developing a separate process to decide, for each utterance, whether to ask a question.

In this section, we describe the two methods -- which we call Conditional Training (CT) and Weighted Decoding (WD) -- that we use to control attributes of the output $y$ at the dialogue level.

\subsection{Conditional Training (CT)}
\label{subsec:ct}
Conditional Training \citep{fan2018controllable, kikuchi2016controlling, peng2018towards} is a method to learn a sequence-to-sequence model $P(y|x,z)$, where $z$ is a discrete \textit{control variable}.
If the control attribute is naturally continuous (for example in our work, repetitiveness, specificity and response-relatedness), we use $z$ to represent bucketed ranges.
For a binary attribute like question-asking, $z$ represents an overall probability (as explained in Section \ref{sec:control_methods}).

To train a CT model, we first automatically annotate every $(x,y)$ pair in the training set with the attribute we wish to control (for example, whether $y$ contains a question mark). During training, for each example we determine the corresponding $z$ value (for continuous attributes, this simply means sorting into the correct bucket; for question-asking, see Section \ref{subsec:qn}).
Next, the control variable $z$ is represented via an embedding (each of the possible values of $z$ has its own embedding).
For all our experiments, the embedding is of length 10; this was determined via hyperparameter tuning.
There are several possible ways to condition the sequence-to-sequence model on $z$ -- for example, append $z$ to the end of the input sequence, or use $z$ as the START symbol for the decoder. 
We find it most effective to concatenate $z$ to the decoder's input on every step.\footnote{To build a CT model $P(y|x,z_1,\dots,z_n)$ conditioned on \textit{multiple} controls $\{z_1,\dots,z_n\}$, we can simply concatenate multiple control embeddings to the decoder inputs.}
Lastly, the CT model learns to produce $y=y_1,\dots,y_T$ by optimizing the cross-entropy loss:
\begin{equation*}
\label{eqn:conditional_loss}
    \text{loss}_{\text{CT}} = - \frac{1}{T} \sum_{t=1}^T \log P(y_t | x, z, y_1, \dots, y_{t-1})
\end{equation*}
Our CT models are initialized with the parameters from the baseline sequence-to-sequence model $P(y|x)$ (the new decoder parameters are initialized with small random values), then fine-tuned to optimize $\text{loss}_{\text{CT}}$ on the PersonaChat training set, until convergence of $\text{loss}_{\text{CT}}$ on the validation set.

\subsection{Weighted Decoding (WD)}
\label{subsec:wd}
Weighted Decoding \citep{ghazvininejad2017hafez} is a decoding method that increases or decreases the probability of words with certain features.
The technique is applied only at test time, requiring no change to the training method. 
A limitation of WD is that the controllable attribute must be defined at the word-level; any desired utterance-level attribute must be redefined via word-level features.

In weighted decoding, on the $t^{th}$ step of decoding, a partial hypothesis $y_{<t}=y_1,\dots,y_{t-1}$ is expanded by computing the score for each possible next word $w$ in the vocabulary:
\begin{align*}
    &\text{score}(w,y_{<t};x) = \text{score}(y_{<t};x)  \\
    &+ \log P_{{\text{RNN}}}(w| y_{<t},x)  + \sum_i w_i * f_i(w;y_{<t},x).
\end{align*}
Here, $\log P_{\text{RNN}}(w| y_{<t},x)$ is the log-probability of the word $w$ calculated by the RNN, $\text{score}(y_{<t};x)$ is the accumulated score of the already-generated words in the hypothesis $y_{<t}$, and $f_i(w;y_{<t},x)$ are \textit{decoding features} with associated weights $w_i$.
There can be multiple features $f_i$ (to control multiple attributes), and the weights $w_i$ are hyperparameters to be chosen.

A decoding feature $f_i(w;y_{<t},x)$ assigns a real value to the word $w$, in the context of the text generated so far $y_{<t}$ and the context $x$.
The feature can be continuous (e.g. the unigram probability of $w$), discrete (e.g. the length of $w$ in characters), or binary (e.g. whether $w$ starts with the same letter as the last word in $y_{<t}$). A positive weight $w_i$ increases the probability of words $w$ that score highly with respect to $f_i$; a negative weight decreases their probability.

Note that weighted decoding and conditional training can be applied simultaneously (i.e. train a CT model then apply WD at test time) -- a strategy we use in our experiments.

\section{Controlling conversational attributes}
\label{sec:auto_eval}
In this section, we describe how we use conditional training and weighted decoding to control four attributes: repetition, specificity, response-relatedness and question-asking.
We evaluate the effectiveness of both control methods via automatic metrics (i.e., measuring how well the attribute was controlled), and use our findings to select control methods and control settings to be explored further via human evaluation (Section \ref{sec:human_eval}).

\subsection{Repetition}
\label{subsec:rep}
Our baseline model exhibits three types of repetition, which we call \textit{external repetition} (self-repetition across utterances), \textit{internal repetition} (self-repetition within utterances), and \textit{partner repetition} (repeating the conversational partner).

To control repetition with weighted decoding,\footnote{We also tried controlling repetition with conditional training, defining $z$ as the (bucketed) maximum ROUGE-L precision between the response $y$ and the bot's previous utterances.
However, this method was unsuccessful
because there are not enough repetitive examples in the training data for the model to learn the control.
Experimenting with data augmentation to solve this problem is an area for future work.
} we define five $n$-gram based decoding features (see Appendix \ref{appendix:rep_feats}).
Three of these features ({\smalltt{extrep\_bigram}}, {\smalltt{intrep\_bigram}} and {\smalltt{partnerrep\_bigram}}) identify repeating bigrams for the three repetition types. 
The other two features (\smalltt{extrep\_unigram} and \smalltt{intrep\_unigram}) identify repeating content words.
By applying a negative weight to these features, we can reduce repetition.
In particular, if the weight is $-\infty$, our method is equivalent to \textit{n-gram blocking} as described by \cite{kulikov2018importance}.
We observe that repetition control is very important, thus all further control experiments include repetition control.

\subsection{Specificity}
\label{subsec:spec}

Like many sequence-to-sequence models using beam search decoding, our baseline frequently asks generic questions such as \textit{What music do you like?} and gives dull, unspecific responses, such as \textit{I like all kinds of music}.

We control specificity using Normalized Inverse Document Frequency (NIDF) as a measure of word rareness.\footnote{Note that our NIDF specificity features are similar to the NIRF and NIWF features used by \cite{zhang2018learning}.}
The Inverse Document Frequency of a word $w$ is $\text{IDF}(w) = \log(R/c_w)$ where $R$ is the number of responses in the dataset, and $c_w$ is the number of those responses that contain $w$.
Normalized IDF (which ranges from 0 to 1) is 
\begin{align}
    \text{NIDF}(w) = \frac{\text{IDF}(w) - \smalltt{min\_idf}}{\smalltt{max\_idf} - \smalltt{min\_idf}}
\end{align}
where $\smalltt{min\_idf}$ and $\smalltt{max\_idf}$ are the minimum and maximum IDFs, taken over all words in the vocabulary.
To control specificity with weighted decoding, we use NIDF as a decoding feature.
As shown in Table \ref{tab:spec_ex}, this method produces reasonable outputs when the feature weight is within a certain range, but at the extremes the outputs are nonsensical.
The boundary for nonsensical output differs from example to example.

To control specificity with conditional training, we define the specificity of an utterance $y$ to be the mean NIDF of the words in $y$.
Thus our control variable $z$ is mean NIDF (discretized into 10 equal-sized buckets).
As shown in Table \ref{tab:spec_ex}, this method gives outputs with a narrower NIDF range, but overall produces less nonsensical outputs.

\begin{table}[t]
\begin{center}
\small
\begin{tabular}{lrl}
    \toprule
    \multicolumn{3}{l}{\textbf{Input:} \textit{Yes, I'm studying law at the moment}}\\ 
    \multicolumn{3}{l}{\textbf{Baseline Response:} \textit{That sounds like a lot of fun!}}\\
    \midrule
    \textbf{Wt} & \textbf{NIDF} & \textbf{Weighted Decoding Response} \\
    -5.0 & 0.6\% & \textit{Oh......................................}\\
    0.0 & 17.1\% & \textit{That sounds like a lot of fun!} \\ 
    3.0 & 18.3\% & \textit{That sounds like a lot of fun. How} \\
    & & \textit{long have you been studying?} \\  
    7.0 & 38.5\% & \textit{I majored in practising my} \\ 
    & & \textit{spiritual full time philosophy test} \\ 
    10.0 & 71.9\% & \textit{Oh wow! Merna jean isa paino yi} \\
     & & \textit{hao hui bu acara sya gila} [\ldots] \\
    \midrule
    $z$ & \textbf{NIDF} & \textbf{Conditional Training Response} \\
    0 & 16.8\% & \textit{Sounds like you are a great person!} \\
    2 & 18.3\% & \textit{So you are a law student?} \\ 
    4 & 18.4\% & \textit{That sounds like a lot of fun} \\
    6 & 22.8\% & \textit{That sounds like a rewarding job!} \\
    8 & 24.4\% & \textit{That sounds like a rewarding career!} \\
    \bottomrule
\end{tabular}
\end{center}
\caption{Middle: Example of controlling specificity (NIDF) via weighted decoding. At the extremes, the model produces only the most rare or the most common tokens. Bottom: Example of controlling specificity via conditional training. This gives a narrower NIDF range, but all the responses are appropriate.}
\label{tab:spec_ex}
\end{table}

\subsection{Response-relatedness}
\label{subsec:resp}
In conversation, it's generally desirable to produce a response that is related to the partner's last utterance; for example if the partner says \textit{My grandfather died last month}, it is appropriate to say \textit{I'm so sorry. Were you close to your grandfather?} 
However, our baseline model frequently responds with unrelated utterances like \textit{Do you have any pets?}

To control response-relatedness with weighted decoding, we use the decoding feature \smalltt{resp\_rel}:
\begin{align*}
\smalltt{resp\_rel}(w;y_{<t},x) &= \\ \smalltt{cos\_sim}&(\smalltt{word\_emb}(w),\smalltt{sent\_emb}(\ell))
\end{align*}
where $\smalltt{word\_emb}(w)$ is the GloVe embedding for the word $w$, $\smalltt{sent\_emb}(\ell)$ is the sentence embedding for the partner's last utterance $\ell$ (note $\ell$ is part of the context $x$), and $\smalltt{cos\_sim}$ is the cosine similarity between the two.
In particular, the sentence embedding $\smalltt{sent\_emb}(s)$ for an utterance $s$ is a weighted average of the GloVe embeddings of the words in $s$, with the first principal component projected out; for full details, see \cite{arora2017simple}.
This method of controlling response-relatedness is similar to that described in \citep{baheti2018generating}.
We find that weighted decoding is effective to control the semantic relatedness of the model's response to the partner's last utterance (see Table \ref{tab:resp_wd_ex}). As before, we find that extreme weights lead to nonsensical output.

To control response-relatedness with conditional training, we try defining the control variable $z$ to be  $\smalltt{cos\_sim}(\smalltt{sent\_emb}(y),\smalltt{sent\_emb}(\ell))$, the overall cosine similarity between the partner's last utterance $\ell$ and the model's response $y$ (again, we discretize $z$).
However, we find this method ineffective -- the CT model learns only a very weak connection between $z$ and the semantic relatedness of the output (see Section \ref{sec:discuss_methods} for more details).

\begin{table}[t]
\begin{center}
\small
\setlength{\tabcolsep}{4pt} 
\begin{tabular}{rrl}
    \toprule
    \multicolumn{3}{l}{\textbf{Input:} \textit{Do you go get coffee often}}\\ 
    \multicolumn{3}{l}{\textbf{Baseline Response:} \textit{I do, when I am not playing the piano.}}\\
    \midrule
    \textbf{Wt} & \textbf{Sim} & \textbf{Weighted Decoding Response} \\
    -10.0 & -0.05 & \textit{I am a musician.} \\
      0.0 & -0.02 & \textit{I do, when I am not playing the piano.} \\
      5.0 & 0.40 & \textit{I do, usually at starbucks.} \\
      8.0 & 0.59 & \textit{Not usually, especially when you drink latte.} \\ 
     11.0 & 0.72 & \textit{Not often, usually with drinks,} \\ 
     & & \textit{espresso, latte, tea, etc.} \\
    \bottomrule
\end{tabular}
\end{center}
\caption{Example of controlling response-relatedness (cosine similarity to input) via weighted decoding. Positive weights (e.g. 5.0) can yield more on-topic responses, but higher weights (e.g. 11.0) can result in nonsensical lists of topically related words.}
\label{tab:resp_wd_ex}
\end{table}

\subsection{Question-asking}
\label{subsec:qn}
Considerate chitchat requires a reciprocal asking and answering of questions -- asking too few or too many can appear self-centered or nosy. We control question-asking in order to study these trade-offs.

To control question-asking with weighted decoding, 
we use the binary decoding feature $\smalltt{is\_qn\_word}(w)$, which is equal to 1 if and only if the word $w$ is in a pre-defined list of interrogative words (\textit{how, what, when, where, which, who, whom, whose, why, ?}).
We find this is a somewhat effective method to encourage or discourage questions, but with unintended side-effects:
a negative weight can discourage valid non-question utterances that happen to contain interrogative words (such as \textit{I'm learning \textbf{how} to knit}) and a positive weight can result in degenerate utterances (such as \textit{What???????} or \textit{Who? When? How?}).

For conditional training, 
we regard an utterance $y$ as containing a question if and only if $y$ contains a question mark. 
We train our CT model on a control variable $z$ with 11 possible values: $\{0,\dots,10\}$.
As discussed in Section \ref{sec:control_methods}, we wish to control question-asking at the distributional, dialogue level, rather than at the binary, utterance level.
Thus the setting $z=i$ means that the model should produce, on average, utterances containing `?' with probability $i/10$. During training we randomly assign examples to buckets such that each bucket $i$ is trained on examples with the correct proportion of questions ($i/10$), and all buckets have the same amount of training examples.

\begin{figure}[t]
    \centering
    \includegraphics[width=\columnwidth]{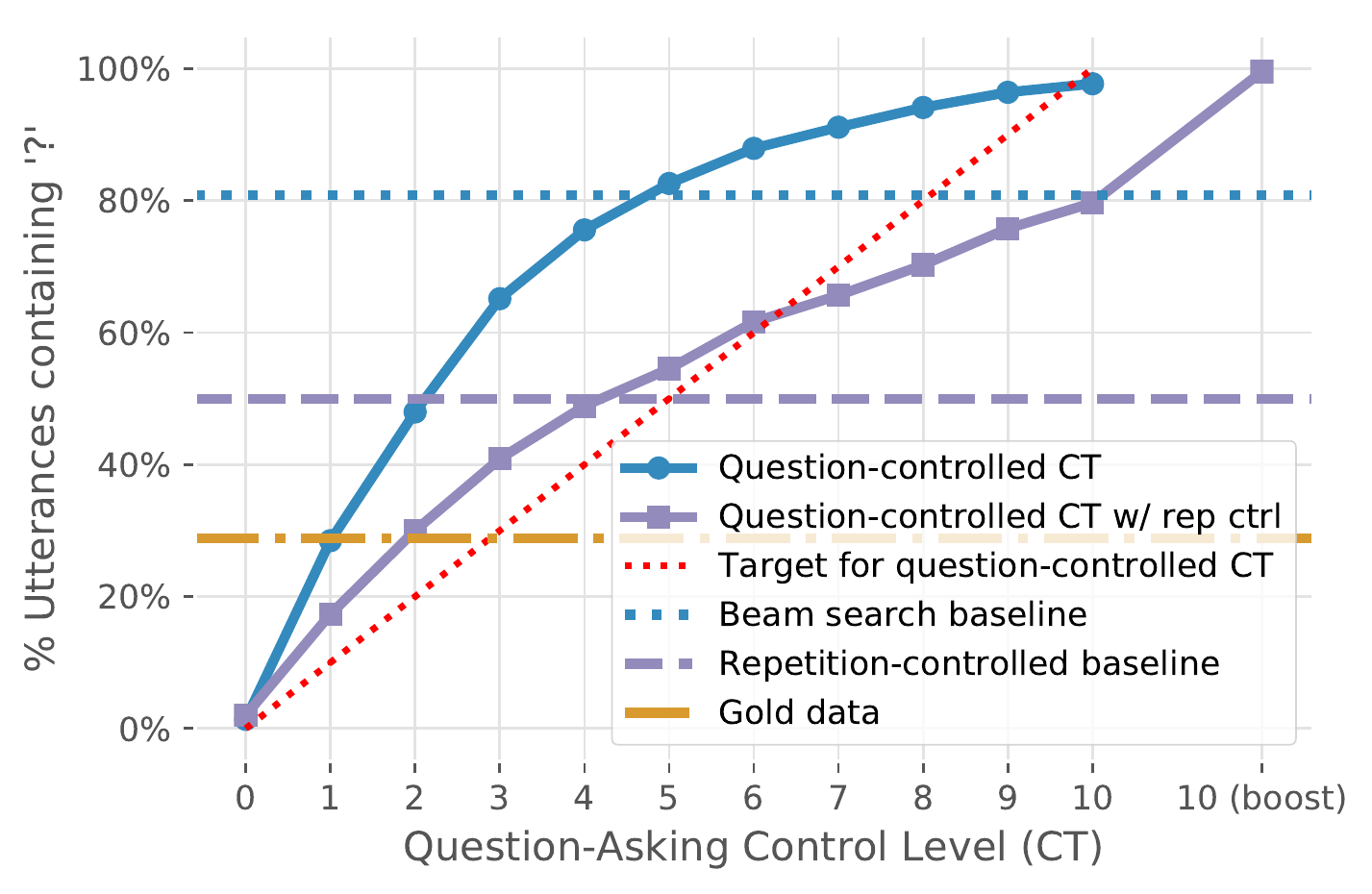}
    \caption{Controlling question-asking via Conditional Training (CT). See Appendix \ref{appendix:auto_metrics} for exact numbers.}
    \label{fig:qn_faithfulness}
\end{figure}

We find that conditional training is effective to control question-asking -- as shown in Figure \ref{fig:qn_faithfulness}, by increasing $z$ from 0 to 10, we obtain a range of question-asking rates from 1.40\% to 97.72\%. 
However, when we introduce repetition control, question-asking is reduced -- in particular, the $z=10$ setting (which should produce 100\% questions) now only produces 79.67\% questions. 
The primary problem is the weighted decoding feature \smalltt{extrep\_bigram}, which discourages bigrams that have appeared in previous utterances -- this prevents the model from producing bigrams that commonly occur in many questions, such as \textit{do you} and \textit{what is}. 
To fix this, we introduce an extra setting $z=10$ \textit{(boost)}, in which we do not use the feature \smalltt{extrep\_bigram} for weighted decoding during beam search, but we do use it to rerank the candidates after beam search.
This setting, which allows the model to produce necessary question-asking bigrams, yields a 99.54\% question-asking rate, at the cost of slightly increased external bigram repetition (see Appendix \ref{appendix:auto_metrics}).

For controlling question-asking, conditional training is preferable to weighted decoding for two reasons. 
Firstly, it allows us to achieve (close to) 0\% questions, 100\% questions, or anything in between, without introducing the risk of degenerate output. 
Secondly, presence-of-a-question-mark captures the true attribute of interest (question-asking) more exactly and directly than presence of interrogative words.
For these reasons, only the CT method is considered in the human evaluation.

\section{Comparison of control methods}
\label{sec:discuss_methods}
The previous section shows that conditional training and weighted decoding are both useful techniques, with different strengths and weaknesses. 

The primary disadvantage of conditional training is that it sometimes fails to learn the connection between the control variable $z$ and the target output $y$.
In practice, we find the model can learn simple attributes of the output (such as the presence of `?', and overall genericness), but not relationships between the input and output (such as semantic relatedness).
By contrast, weighted decoding can force the desired feature to appear in the output by raising the weight arbitrarily high (though this may have unintended side-effects).


The primary disadvantage of weighted decoding is that it risks going off-distribution when the weight is too strong.
By contrast, conditional training produces mostly well-formed, in-distribution outputs.
This highlights the importance of learned control -- it is safer to learn to produce output that both satisfies the control variable and is appropriate, than to alter the decoding process to satisfy the control variable, potentially trading off appropriateness in the process.

Other considerations include:
(1) Convenience: conditional training requires retraining; weighted decoding doesn't, but is slower at test time.
(2) Data availability: conditional training requires training examples of the controllable attribute, whereas weighted decoding can control any computable feature without requiring examples.
(3) Attribute definition: conditional training can control sentence-level attributes, but they must be discrete. By contrast, weighted decoding requires word-level features, but they can be continuous.

\section{Human evaluation results}
\label{sec:human_eval}
In order to study the effect of our controllable attributes, we conduct a large-scale human evaluation of 28 model configurations (see Appendix \ref{appendix:control_settings}), plus human-human conversations for comparison.

\paragraph{Approach}
In our evaluation, a crowdworker chats with a model (or in the human-human case, another crowdworker) for six conversational turns, then answers eight multiple-choice questions which each capture different aspects of conversational quality: avoiding repetition, interestingness, making sense, fluency, listening, inquisitiveness, humanness and engagingness. 
The eight questions are Likert questions on a 1-4 scale, where higher is better.\footnote{Exceptions: Avoiding repetition is a 1-3 scale, as we found this gave clearer instructions. Inquisitiveness has an optimal score of 3; 1 and 2 represent too little question-asking, and 4 represents too much.}
To match the ConvAI2 Challenge, we also add a persona retrieval question, in which the crowdworker is asked to select which of two possible personas was the model's persona.
For full details of the evaluation design, see Appendix \ref{appendix:human_eval_questions}.

Our evaluation is the same as the ConvAI2 Challenge evaluation, but more detailed -- ConvAI2 includes only engagingness and persona retrieval.\footnote{There are three other minor differences between our evaluation and ConvAI2's: (1) We fix capitalization and spacing before showing the chatbot's utterances to crowdworkers, while ConvAI2 show the raw lowercase tokenized form. We found the latter interferes with fluency evaluation. (2) We conduct 6 dialogue turns, while ConvAI2 conducts 4-6. This was necessary to evaluate repetitiveness. 
(3) We use (publicly-available) validation set personas, while ConvAI2 uses (hidden) test set personas. This enables us to release our evaluation chatlogs.}
As in the ConvAI2 challenge, each of our 28 model configurations was evaluated by over 100 crowdworkers,
and the results were adjusted for annotator variance via a Bayesian calibration \citep{kulikov2018importance}.

In designing our evaluation, we aimed to capture the four aspects we expected to directly improve via control (avoiding repetition, interestingness, listening, inquisitiveness), two important error classes we thought would be affected by our controls (fluency, making sense), and two overall quality measures (engagingness, humanness).

\begin{figure*}[t]
\minipage{0.33\textwidth}%
  \centering
    \includegraphics[width=\columnwidth]{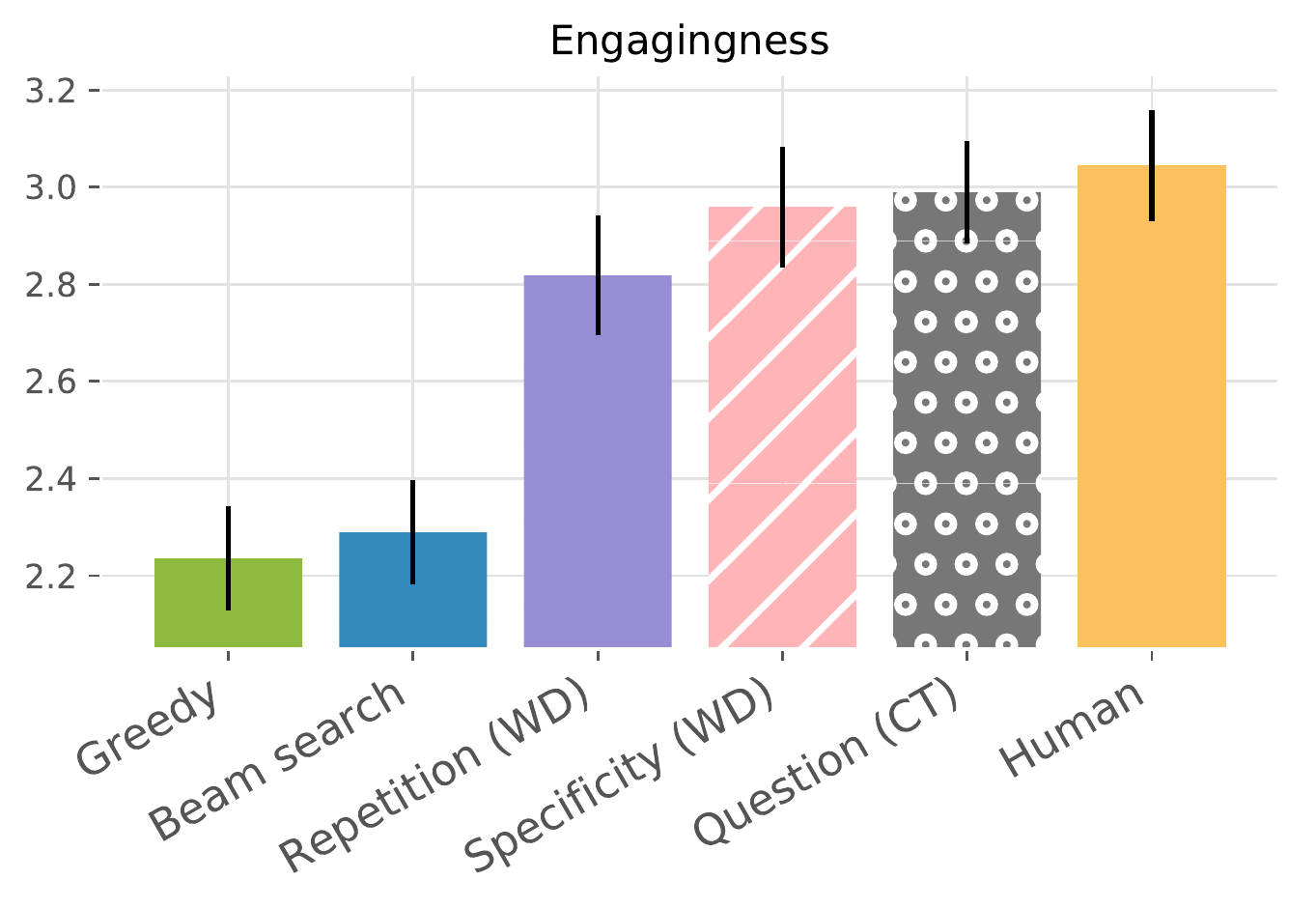}
\endminipage\hfill
\minipage{0.33\textwidth}
    \centering
    \includegraphics[width=\columnwidth]{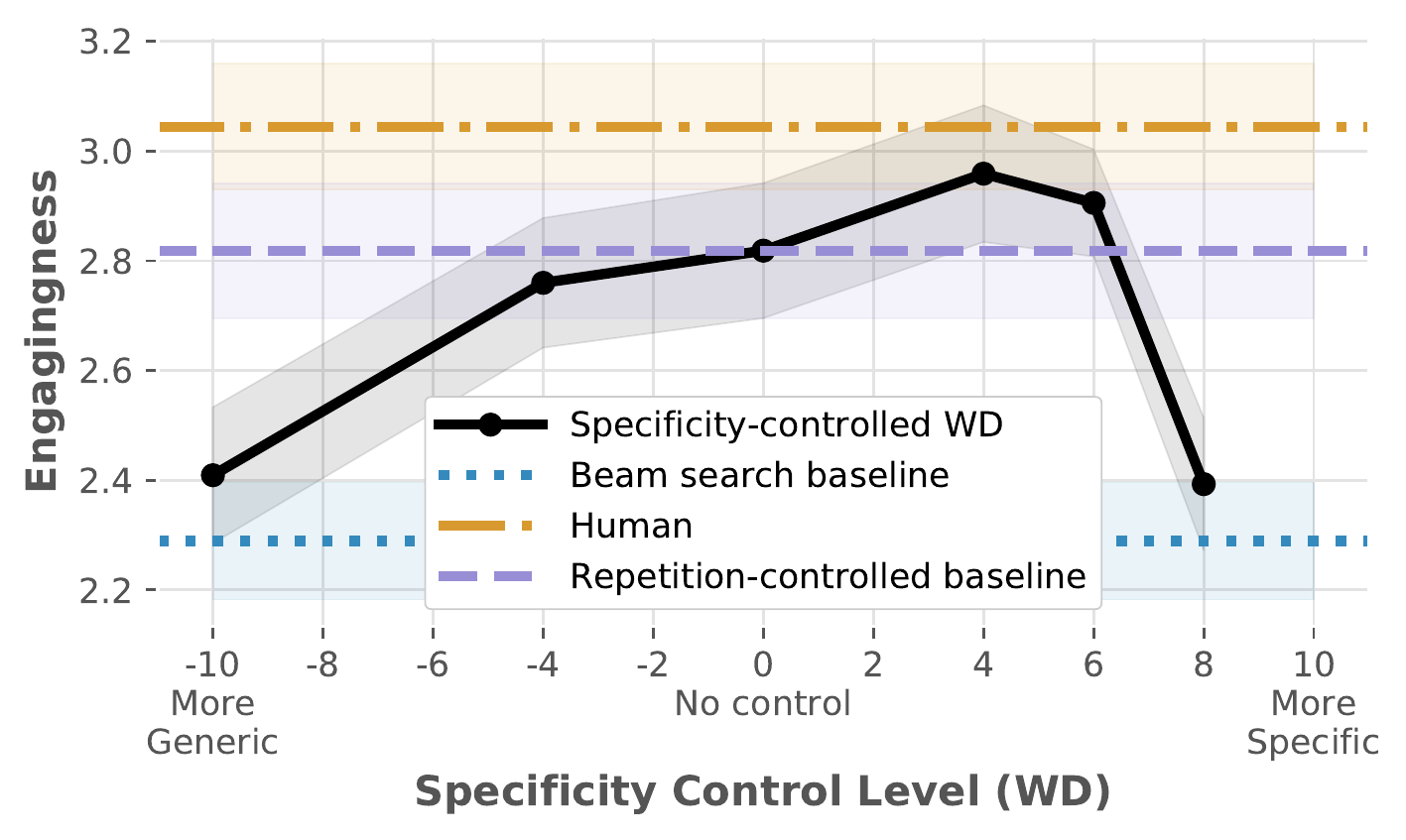}
\endminipage\hfill
\minipage{0.33\textwidth}
    \centering
    \includegraphics[width=\columnwidth]{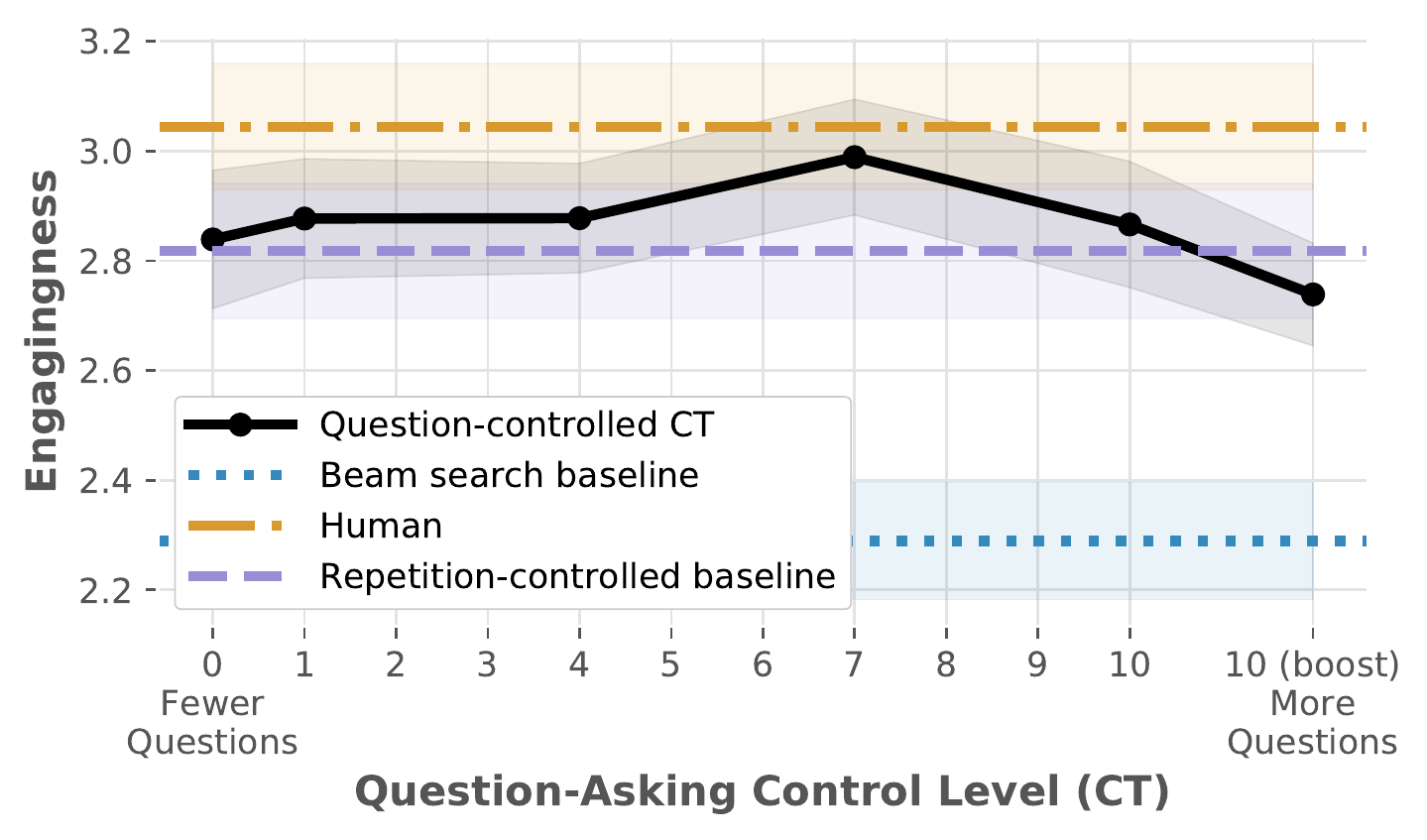}
\endminipage\hfill
\vspace{-10pt}
\caption{Calibrated human judgments of engagingness for the baselines and best controlled models (left); for different specificity control settings (middle); and for different question-asking control settings (right).
    }
\label{fig:engage_results}
\end{figure*}

\begin{figure*}[t]
    \centering
    \includegraphics[width=\textwidth]{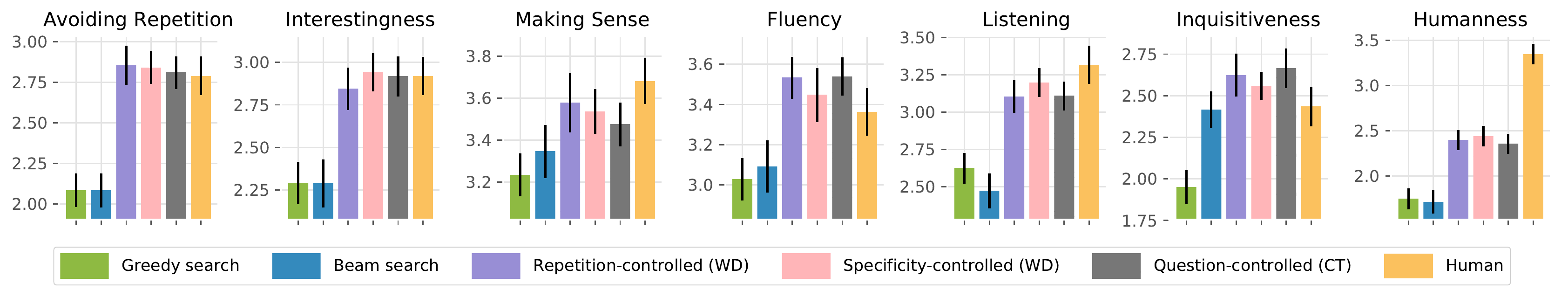}
    \caption{
    Calibrated human judgments of conversational aspects for the baselines and best controlled models. 
    Note: In Figure \ref{fig:engage_results} and here, the Specificity and Question controlled models both include Repetition control, but Question control doesn't include Specificity control, or vice versa. See Table \ref{tab:calibrated} for exact numbers.}
    \label{fig:birthday}
\end{figure*}

\begin{table*}[t]
\footnotesize
\begin{center}
\begin{tabular}{lrl}
\toprule
\textbf{Model} & \textbf{Win\%} & \textbf{Top 3 reasons for preferring model} \\
\midrule
Specificity WD ($\text{weight}=6$) & 84.1\% &  \textit{More information; Better flow; More descriptive} \\
Specificity WD ($\text{weight}=4$) & 75.5\% &  \textit{More information; They describe their life in more detail; Funny} \\ 
Specificity CT ($z=7$) & 56.2\% &  \textit{More information; Better flow; Seems more interested} \\
\bottomrule
\end{tabular}
\end{center}
\caption{A/B tests comparing various specificity-controlled models to the repetition-controlled baseline on interestingness. We find all comparisons are significant ($p<.05$; binomial test).
}
\label{tab:ab_results}
\end{table*}

\subsection{Main findings}
\label{subsec:main_findings}
In this section we summarize the main findings of our human evaluation -- whose full results can be found in Appendices \ref{appendix:human_eval_table} and \ref{appendix:human_eval_plots}, with sample conversations in Appendix \ref{appendix:conversation_examples}.

As Figure \ref{fig:engage_results} shows, controlling for repetition, specificity and question-asking all lead to large engagingness improvements over the greedy and beam-search baseline models.
In particular, we find that controlling for multi-turn (self) repetition is important and should be incorporated alongside other attribute control methods. 
We found no improvement by controlling response-relatedness.

To better understand these overall engagingness improvements, we consider the full set of human judgments, shown in Figure \ref{fig:birthday}. 
We find that reducing repetition leads to improvements across all our aspects of conversational quality.
Increasing specificity shows improvements in interestingness and listening ability over the repetition-controlled baseline, while
increasing question-asking shows improvements in inquisitiveness and interestingness over the repetition-controlled baseline.

Our most engaging model, which controls both repetition and question-asking -- marked `Question (CT)' in Figure \ref{fig:engage_results} (left) -- matches the engagingness of the winning entry in the ConvAI2 competition, as both models achieve a raw score\footnote{Although the same Bayesian calibration method was applied both in our study and in the ConvAI2 competition, calibrated scores are not comparable across the two; thus we compare raw scores (viewable in Table \ref{tab:raw}).} of $3.1$ \citep{dinan2019second}.
However, the ConvAI2 winner, Lost in Conversation, was trained on approximately 12$\times$ as much data as our model.
Lost in Conversation is based on the OpenAI GPT Language Model \citep{radford2018improving}, which is pretrained on the BookCorpus \citep{zhu2015aligning}, which contains approximately 985 million words, whereas our model is pretrained on the Twitter dataset (approximately 79 million words).

Altogether, our evaluation clearly shows that controlling low-level attributes over multiple turns leads to improved overall quality.

\subsection{Effect of controlled attributes}
\label{subsec:controlled_attributes_results}

\paragraph{Repetition (WD)}
We observe that self-repetition across utterances (\textit{external repetition}) is by far the most severe form of repetition in our beam search baseline model.
We evaluate several settings of the \smalltt{extrep\_bigram} weighted decoding feature, and find that an aggressive repetition-reduction setting (reducing bigram repetition rate to below gold data levels) is rated best.
We also find that blocking repeated content words improves the avoiding repetition score. See Appendices \ref{appendix:control_settings}, \ref{appendix:auto_metrics} and \ref{appendix:human_eval_table} for full details.

As shown in Figure \ref{fig:engage_results} (left) and Figure \ref{fig:birthday}, our repetition-controlled model improves hugely over the beam search baseline in all metrics, and achieves close-to-human scores on all metrics except humanness.
This striking result demonstrates that repetition is by far the biggest limiting quality factor for naive sequence-to-sequence dialogue agents. 
The result also emphasizes the importance of \textit{multi-turn} dialogue evaluation to detect the problem.
We refer to this model as the \textit{repetition-controlled baseline}, and use it as a basis for all remaining experiments (i.e., we control specificity, response-relatedness and question-asking on top of these repetition-control settings).

\paragraph{Specificity (WD, CT)}
For our weighted decoding models, the extreme settings (very generic and very specific) score poorly in engagingness due to the frequent presence of degenerate output -- see Figure \ref{fig:engage_results} (middle).
We find that the $\text{weight}=4$ setting (which is more specific than the repetition-controlled baseline and about as specific as the gold data) maximizes engagingness.
As shown in Figure \ref{fig:engage_results} (left) and Figure \ref{fig:birthday}, this more-specific model is rated more interesting, engaging, and a better listener than the repetition-controlled baseline, but at the cost of reduced fluency and making sense.
Our CT model with $z=7$ (which has a similar NIDF level as WD with $\text{weight}=4$) shows similar results, but the improvements are smaller.
For further discussion on the interestingness of our specificity models, see Section \ref{subsec:ab_eval}.

\paragraph{Response-relatedness (WD)}
We evaluated several control settings ($\text{weight}=$ $-10,5,10,13$) and found that none scored better than $\text{weight}=0$ (no response-relatedness control); see Appendix \ref{appendix:human_eval_plots}.
This is surprising -- prior to running the human evaluation, we annotated 100 examples ourselves to determine the best control settings.
While we identified a more responsive setting ($\text{weight}=5$) as less likely than the uncontrolled model to ignore the user, crowdworkers rated it as a slightly \textit{worse} listener than the uncontrolled model.
One explanation for this discrepancy is that the more responsive model takes more risks, using more rare words (0.197 NIDF, up from 0.178), and thus receives a lower makes-sense score (3.41, down from 3.70).
We hypothesize that, compared to us, the crowdworkers are less tolerant of slightly nonsensical output, and more tolerant of generic unrelated utterances.

\paragraph{Question-asking (CT)}
As shown in Figure \ref{fig:engage_results} (right), a question-asking rate of 65.7\% ($z=7$) maximizes engagingness. 
This setting, which asks more questions than both the repetition-controlled baseline (50.0\%) and the human-produced gold data (28.8\%), brings us closest to human-level engagingness -- see Figure \ref{fig:engage_results} (left).
Although we find that a rate of approximately 65.7\% question-asking is the most engaging, a lower level (48.9\%, or $z=4$) is rated the best listener.
Lastly, we find that although asking too many questions is less engaging, most crowdworkers will not directly criticize a chatbot that asks questions on every turn -- only 11.9\% of crowdworkers judged the $z=10$ \textit{(boost)} setting, which asks 99.5\% questions, as asking too many questions.\footnote{Though this conclusion may hold true for the PersonaChat task -- a synthetic chatting task that instructs participants to get to know each other -- in real-life social conversations, incessant question-asking may be less tolerated.}
For full details of these scores, see Appendix \ref{appendix:auto_metrics} and \ref{appendix:human_eval_plots}.

For time and budget reasons, we did not evaluate any models controlling both question-asking and specificity. However, we expect it is possible to obtain further improvements by doing so.

\subsection{A/B tests for interestingness}
\label{subsec:ab_eval}
Though our more-specific models yielded significant improvements in engagingness, we were surprised that they did not yield clearer improvements in interestingness.
To investigate further, we conducted an A/B interestingness evaluation of three specificity-controlled models, compared to the repetition-controlled baseline.
Crowdworkers were shown two conversations (from the main human evaluation) and asked to choose which model was more interesting (see Figure \ref{fig:ab_ui} for details). We collected 500 samples per comparison, plus 200 additional human vs repetition-controlled baseline samples, which were used to filter for quality control. After discarding low-quality crowdworkers, we have roughly 300 evaluations per comparison, with an average Cohen's $\kappa=0.6$.

As shown in Table \ref{tab:ab_results}, all three models were rated significantly more interesting than the repetition-controlled baseline.
This convincingly shows that producing utterances with more rare words is a valid strategy to improve interestingness.
We have two explanations for why these interestingness differences did not materialize in our main evaluation.
Firstly, interestingness is a particularly subjective metric (unlike more tangible metrics such as avoiding repetition and making sense) -- this makes it hard to calibrate across crowdworkers.
Secondly, we suspect that in our original evaluation, the crowdworkers may have evaluated the interestingness of the \textit{task} rather than the \textit{chatbot}. 
This could account for why subtle increases in conversational ability did not result in higher interestingness ratings -- the PersonaChat task itself has a natural interestingness limit.

\section{Conclusion}
\label{sec:conclusion}

\paragraph{What makes a good conversation?}
Through our evaluation, we showed that a good conversation is about balance -- controlling for the right level of repetition, specificity and question-asking is important for overall quality.
We also found that conversational aspects such as interestingness, listening, and inquisitiveness are all important -- though optimizing these can introduce a trade-off against certain types of errors (such as repetitive, disfluent, or nonsensical output).
Secondly, multi-turn evaluation is essential to study what makes a good conversation -- multiple turns are required to reveal issues such as repetition, consistency, and question-asking frequency.
Lastly, what do we mean by `good'?
Although humanness and engagingness are both commonly used as overall quality metrics, the two are very different.
While our models achieved close-to-human scores on engagingness, they failed to get close on humanness -- showing that a chatbot need not be human-like to be enjoyable.
This striking result also demonstrates the importance of measuring more than one quality metric when evaluating dialogue agents.
  
\paragraph{Outlook}
Our work shows that neural generative systems have systemic problems when applied to open-ended dialogue, some of which (e.g. repetition) are only observable in the multi-turn setting. 
Furthermore, control of low-level attributes offers a practical way to correct these problems, yielding large improvements to overall quality -- in our case, comparable to systems trained on much more data.
Future work includes optimizing control settings automatically, and building more convincingly human-like chatbots.

\newpage
\bibliography{main}

\begin{thebibliography}{41}
\expandafter\ifx\csname natexlab\endcsname\relax\def\natexlab#1{#1}\fi

\bibitem[{Arora et~al.(2017)Arora, Liang, and Ma}]{arora2017simple}
Sanjeev Arora, Yingyu Liang, and Tengyu Ma. 2017.
\newblock \href {https://openreview.net/pdf?id=SyK00v5xx} {A simple but
  tough-to-beat baseline for sentence embeddings}.
\newblock In \emph{Proceedings of the International Conference on Learning
  Representations (ICLR)}.

\bibitem[{Baheti et~al.(2018)Baheti, Ritter, Li, and
  Dolan}]{baheti2018generating}
Ashutosh Baheti, Alan Ritter, Jiwei Li, and Bill Dolan. 2018.
\newblock \href {http://aclweb.org/anthology/D18-1431} {Generating more
  interesting responses in neural conversation models with distributional
  constraints}.
\newblock In \emph{Proceedings of the 2018 Conference on Empirical Methods in
  Natural Language Processing}, pages 3970--3980. Association for Computational
  Linguistics.

\bibitem[{Bordes et~al.(2017)Bordes, Boureau, and Weston}]{bordes2016learning}
Antoine Bordes, Y-Lan Boureau, and Jason Weston. 2017.
\newblock \href {https://arxiv.org/pdf/1605.07683.pdf} {Learning end-to-end
  goal-oriented dialog}.
\newblock In \emph{Proceedings of the International Conference on Learning
  Representations (ICLR)}.

\bibitem[{Dinan et~al.(2019)Dinan, Logacheva, Malykh, Miller, Shuster, Urbanek,
  Kiela, Szlam, Serban, Lowe et~al.}]{dinan2019second}
Emily Dinan, Varvara Logacheva, Valentin Malykh, Alexander Miller, Kurt
  Shuster, Jack Urbanek, Douwe Kiela, Arthur Szlam, Iulian Serban, Ryan Lowe,
  et~al. 2019.
\newblock \href {https://arxiv.org/pdf/1902.00098.pdf} {The second
  conversational intelligence challenge (convai2)}.
\newblock \emph{arXiv preprint arXiv:1902.00098}.

\bibitem[{Dinan et~al.(2018)Dinan, Roller, Shuster, Fan, Auli, and
  Weston}]{dinan2018wizard}
Emily Dinan, Stephen Roller, Kurt Shuster, Angela Fan, Michael Auli, and Jason
  Weston. 2018.
\newblock \href {https://arxiv.org/pdf/1811.01241.pdf} {Wizard of {W}ikipedia:
  Knowledge-powered conversational agents}.
\newblock \emph{arXiv preprint arXiv:1811.01241}.

\bibitem[{El~Asri et~al.(2017)El~Asri, Schulz, Sharma, Zumer, Harris, Fine,
  Mehrotra, and Suleman}]{asri2017frames}
Layla El~Asri, Hannes Schulz, Shikhar Sharma, Jeremie Zumer, Justin Harris,
  Emery Fine, Rahul Mehrotra, and Kaheer Suleman. 2017.
\newblock \href {http://aclweb.org/anthology/W17-5526} {Frames: a corpus for
  adding memory to goal-oriented dialogue systems}.
\newblock In \emph{Proceedings of the 18th Annual SIGDIAL Meeting on Discourse
  and Dialogue}, pages 207--219, Saarbr\"ucken, Germany. Association for
  Computational Linguistics.

\bibitem[{Fan et~al.(2018)Fan, Grangier, and Auli}]{fan2018controllable}
Angela Fan, David Grangier, and Michael Auli. 2018.
\newblock \href {http://aclweb.org/anthology/W18-2706} {Controllable
  abstractive summarization}.
\newblock In \emph{Proceedings of the 2nd Workshop on Neural Machine
  Translation and Generation}, pages 45--54. Association for Computational
  Linguistics.

\bibitem[{Ficler and Goldberg(2017)}]{ficler2017controlling}
Jessica Ficler and Yoav Goldberg. 2017.
\newblock \href {https://doi.org/10.18653/v1/W17-4912} {Controlling linguistic
  style aspects in neural language generation}.
\newblock In \emph{Proceedings of the Workshop on Stylistic Variation}, pages
  94--104. Association for Computational Linguistics.

\bibitem[{Ghazvininejad et~al.(2017)Ghazvininejad, Shi, Priyadarshi, and
  Knight}]{ghazvininejad2017hafez}
Marjan Ghazvininejad, Xing Shi, Jay Priyadarshi, and Kevin Knight. 2017.
\newblock \href {http://aclweb.org/anthology/P17-4008} {Hafez: an interactive
  poetry generation system}.
\newblock In \emph{Proceedings of ACL 2017, System Demonstrations}, pages
  43--48. Association for Computational Linguistics.

\bibitem[{Guo et~al.(2018)Guo, Metallinou, Khatri, Raju, Venkatesh, and
  Ram}]{guo2018topic}
Fenfei Guo, Angeliki Metallinou, Chandra Khatri, Anirudh Raju, Anu Venkatesh,
  and Ashwin Ram. 2018.
\newblock \href {https://arxiv.org/pdf/1801.03622.pdf} {Topic-based evaluation
  for conversational bots}.
\newblock \emph{Advances in Neural Information Processing Systems,
  Conversational AI Workshop}.

\bibitem[{Hastie(2012)}]{lemon}
Helen Hastie. 2012.
\newblock \href {https://doi.org/10.1007/978-1-4614-4803-7_7} {\emph{Metrics
  and evaluation of spoken dialogue systems}}, pages 131--150. Springer.

\bibitem[{Henderson et~al.(2014)Henderson, Thomson, and
  Williams}]{henderson2014second}
Matthew Henderson, Blaise Thomson, and Jason~D Williams. 2014.
\newblock \href {http://www.aclweb.org/anthology/W14-4337} {The second dialog
  state tracking challenge}.
\newblock In \emph{Proceedings of the 15th Annual Meeting of the Special
  Interest Group on Discourse and Dialogue (SIGDIAL)}, pages 263--272.

\bibitem[{Hu et~al.(2017)Hu, Yang, Liang, Salakhutdinov, and
  Xing}]{hu2017toward}
Zhiting Hu, Zichao Yang, Xiaodan Liang, Ruslan Salakhutdinov, and Eric~P Xing.
  2017.
\newblock \href {https://arxiv.org/pdf/1703.00955.pdf} {Toward controlled
  generation of text}.
\newblock In \emph{Thirty-fourth International Conference on Machine Learning}.

\bibitem[{Kikuchi et~al.(2016)Kikuchi, Neubig, Sasano, Takamura, and
  Okumura}]{kikuchi2016controlling}
Yuta Kikuchi, Graham Neubig, Ryohei Sasano, Hiroya Takamura, and Manabu
  Okumura. 2016.
\newblock \href {https://doi.org/10.18653/v1/D16-1140} {Controlling output
  length in neural encoder-decoders}.
\newblock In \emph{Proceedings of the 2016 Conference on Empirical Methods in
  Natural Language Processing}, pages 1328--1338. Association for Computational
  Linguistics.

\bibitem[{Kulikov et~al.(2018)Kulikov, Miller, Cho, and
  Weston}]{kulikov2018importance}
Ilya Kulikov, Alexander~H Miller, Kyunghyun Cho, and Jason Weston. 2018.
\newblock \href {https://arxiv.org/pdf/1811.00907.pdf} {Importance of a search
  strategy in neural dialogue modelling}.
\newblock \emph{arXiv preprint arXiv:1811.00907}.

\bibitem[{Li et~al.(2016{\natexlab{a}})Li, Galley, Brockett, Gao, and
  Dolan}]{li2015diversity}
Jiwei Li, Michel Galley, Chris Brockett, Jianfeng Gao, and Bill Dolan.
  2016{\natexlab{a}}.
\newblock \href {https://doi.org/10.18653/v1/N16-1014} {A diversity-promoting
  objective function for neural conversation models}.
\newblock In \emph{Proceedings of the 2016 Conference of the North American
  Chapter of the Association for Computational Linguistics: Human Language
  Technologies}, pages 110--119. Association for Computational Linguistics.

\bibitem[{Li et~al.(2017{\natexlab{a}})Li, Monroe, and
  Jurafsky}]{li2017learning}
Jiwei Li, Will Monroe, and Dan Jurafsky. 2017{\natexlab{a}}.
\newblock \href {https://arxiv.org/pdf/1701.06549.pdf} {Learning to decode for
  future success}.
\newblock \emph{arXiv preprint arXiv:1701.06549}.

\bibitem[{Li et~al.(2016{\natexlab{b}})Li, Monroe, Ritter, Jurafsky, Galley,
  and Gao}]{li2016deep}
Jiwei Li, Will Monroe, Alan Ritter, Dan Jurafsky, Michel Galley, and Jianfeng
  Gao. 2016{\natexlab{b}}.
\newblock \href {https://aclweb.org/anthology/D16-1127} {Deep reinforcement
  learning for dialogue generation}.
\newblock In \emph{Proceedings of the 2016 Conference on Empirical Methods in
  Natural Language Processing}, pages 1192--1202, Austin, Texas. Association
  for Computational Linguistics.

\bibitem[{Li et~al.(2017{\natexlab{b}})Li, Monroe, Shi, Jean, Ritter, and
  Jurafsky}]{li2017adversarial}
Jiwei Li, Will Monroe, Tianlin Shi, S{\'e}bastien Jean, Alan Ritter, and Dan
  Jurafsky. 2017{\natexlab{b}}.
\newblock \href {https://arxiv.org/pdf/1701.06547.pdf} {Adversarial learning
  for neural dialogue generation}.
\newblock \emph{arXiv preprint arXiv:1701.06547}.

\bibitem[{Liu et~al.(2016)Liu, Lowe, Serban, Noseworthy, Charlin, and
  Pineau}]{liu2016not}
Chia-Wei Liu, Ryan Lowe, Iulian Serban, Mike Noseworthy, Laurent Charlin, and
  Joelle Pineau. 2016.
\newblock \href {https://aclweb.org/anthology/D16-1230} {How not to evaluate
  your dialogue system: An empirical study of unsupervised evaluation metrics
  for dialogue response generation}.
\newblock pages 2122--2132.

\bibitem[{Lowe et~al.(2017)Lowe, Noseworthy, Serban, Angelard-Gontier, Bengio,
  and Pineau}]{lowe2017towards}
Ryan Lowe, Michael Noseworthy, Iulian~Vlad Serban, Nicolas Angelard-Gontier,
  Yoshua Bengio, and Joelle Pineau. 2017.
\newblock \href {https://doi.org/10.18653/v1/P17-1103} {Towards an automatic
  turing test: Learning to evaluate dialogue responses}.
\newblock In \emph{Proceedings of the 55th Annual Meeting of the Association
  for Computational Linguistics (Volume 1: Long Papers)}, pages 1116--1126.
  Association for Computational Linguistics.

\bibitem[{Lowe et~al.(2015)Lowe, Pow, Serban, and Pineau}]{lowe2015ubuntu}
Ryan Lowe, Nissan Pow, Iulian Serban, and Joelle Pineau. 2015.
\newblock \href {http://aclweb.org/anthology/W15-4640} {The {Ubuntu} dialogue
  corpus: A large dataset for research in unstructured multi-turn dialogue
  systems}.
\newblock In \emph{Proceedings of the 16th Annual Meeting of the Special
  Interest Group on Discourse and Dialogue}, pages 285--294, Prague, Czech
  Republic. Association for Computational Linguistics.

\bibitem[{Miller et~al.(2017)Miller, Feng, Batra, Bordes, Fisch, Lu, Parikh,
  and Weston}]{miller2017parlai}
Alexander Miller, Will Feng, Dhruv Batra, Antoine Bordes, Adam Fisch, Jiasen
  Lu, Devi Parikh, and Jason Weston. 2017.
\newblock \href {http://www.aclweb.org/anthology/D17-2014} {{ParlAI}: A dialog
  research software platform}.
\newblock In \emph{Proceedings of the 2017 Conference on Empirical Methods in
  Natural Language Processing: System Demonstrations}, pages 79--84,
  Copenhagen, Denmark. Association for Computational Linguistics.

\bibitem[{Novikova et~al.(2017)Novikova, Du{\v{s}}ek, Curry, and
  Rieser}]{novikova2017we}
Jekaterina Novikova, Ond{\v{r}}ej Du{\v{s}}ek, Amanda~Cercas Curry, and Verena
  Rieser. 2017.
\newblock \href {https://www.aclweb.org/anthology/D17-1238} {Why we need new
  evaluation metrics for nlg}.
\newblock In \emph{Proceedings of the 2017 Conference on Empirical Methods in
  Natural Language Processing}, pages 2241--2252.

\bibitem[{Parthasarathi and Pineau(2018)}]{parthasarathi2018extending}
Prasanna Parthasarathi and Joelle Pineau. 2018.
\newblock \href {http://www.aclweb.org/anthology/D18-1073} {Extending neural
  generative conversational model using external knowledge sources}.
\newblock In \emph{Proceedings of the 2018 Conference on Empirical Methods in
  Natural Language Processing}, pages 690--695, Brussels, Belgium. Association
  for Computational Linguistics.

\bibitem[{Peng et~al.(2018)Peng, Ghazvininejad, May, and
  Knight}]{peng2018towards}
Nanyun Peng, Marjan Ghazvininejad, Jonathan May, and Kevin Knight. 2018.
\newblock \href {https://doi.org/10.18653/v1/W18-1505} {Towards controllable
  story generation}.
\newblock In \emph{Proceedings of the First Workshop on Storytelling}, pages
  43--49. Association for Computational Linguistics.

\bibitem[{Pennington et~al.(2014)Pennington, Socher, and
  Manning}]{pennington2014glove}
Jeffrey Pennington, Richard Socher, and Christopher Manning. 2014.
\newblock \href {http://www.aclweb.org/anthology/D14-1162} {{GloVe}: Global
  vectors for word representation}.
\newblock In \emph{Proceedings of the 2014 Conference on Empirical Methods in
  Natural Language Processing (EMNLP)}, pages 1532--1543, Doha, Qatar.
  Association for Computational Linguistics.

\bibitem[{Radford et~al.(2018)Radford, Narasimhan, Salimans, and
  Sutskever}]{radford2018improving}
Alec Radford, Karthik Narasimhan, Tim Salimans, and Ilya Sutskever. 2018.
\newblock \href
  {https://pdfs.semanticscholar.org/cd18/800a0fe0b668a1cc19f2ec95b5003d0a5035.pdf}
  {Improving language understanding by generative pre-training}.

\bibitem[{Serban et~al.(2016{\natexlab{a}})Serban, Lowe, Charlin, and
  Pineau}]{serban2016generative}
Iulian~Vlad Serban, Ryan Lowe, Laurent Charlin, and Joelle Pineau.
  2016{\natexlab{a}}.
\newblock \href {https://arxiv.org/pdf/1611.06216.pdf} {Generative deep neural
  networks for dialogue: A short review}.
\newblock \emph{Advances in Neural Information Processing Systems workshop on
  Learning Methods for Dialogue}.

\bibitem[{Serban et~al.(2016{\natexlab{b}})Serban, Sordoni, Bengio, Courville,
  and Pineau}]{serban2016building}
Iulian~Vlad Serban, Alessandro Sordoni, Yoshua Bengio, Aaron~C Courville, and
  Joelle Pineau. 2016{\natexlab{b}}.
\newblock \href {https://dl.acm.org/citation.cfm?id=3016435} {Building
  end-to-end dialogue systems using generative hierarchical neural network
  models}.
\newblock In \emph{AAAI}, volume~16, pages 3776--3784.

\bibitem[{Shen et~al.(2017)Shen, Su, Li, Li, Niu, Zhao, Aizawa, and
  Long}]{shen2017conditional}
Xiaoyu Shen, Hui Su, Yanran Li, Wenjie Li, Shuzi Niu, Yang Zhao, Akiko Aizawa,
  and Guoping Long. 2017.
\newblock \href {https://doi.org/10.18653/v1/P17-2080} {A conditional
  variational framework for dialog generation}.
\newblock In \emph{Proceedings of the 55th Annual Meeting of the Association
  for Computational Linguistics (Volume 2: Short Papers)}, pages 504--509.
  Association for Computational Linguistics.

\bibitem[{Venkatesh et~al.(2017)Venkatesh, Khatri, Ram, Guo, Gabriel, Nagar,
  Prasad, Cheng, Hedayatnia, Metallinou et~al.}]{venkatesh2018evaluating}
Anu Venkatesh, Chandra Khatri, Ashwin Ram, Fenfei Guo, Raefer Gabriel, Ashish
  Nagar, Rohit Prasad, Ming Cheng, Behnam Hedayatnia, Angeliki Metallinou,
  et~al. 2017.
\newblock \href {https://arxiv.org/pdf/1801.03625.pdf} {On evaluating and
  comparing conversational agents}.
\newblock \emph{Advances in Neural Information Processing Systems,
  Conversational AI Workshop}.

\bibitem[{Vinyals and Le(2015)}]{vinyals2015neural}
Oriol Vinyals and Quoc Le. 2015.
\newblock \href {https://arxiv.org/pdf/1506.05869.pdf} {A neural conversational
  model}.
\newblock In \emph{Proceedings of the 31st International Conference on Machine
  Learning, Deep Learning Workshop}, Lille, France.

\bibitem[{Walker et~al.(1997)Walker, Litman, Kamm, and
  Abella}]{walker1997paradise}
Marilyn~A. Walker, Diane~J. Litman, Candace~A. Kamm, and Alicia Abella. 1997.
\newblock \href {https://doi.org/10.3115/976909.979652} {{PARADISE}: A
  framework for evaluating spoken dialogue agents}.
\newblock In \emph{Proceedings of the 35th Annual Meeting of the Association
  for Computational Linguistics}, pages 271--280, Madrid, Spain. Association
  for Computational Linguistics.

\bibitem[{Wang et~al.(2017)Wang, Jojic, Brockett, and
  Nyberg}]{wang2017steering}
Di~Wang, Nebojsa Jojic, Chris Brockett, and Eric Nyberg. 2017.
\newblock \href {https://doi.org/10.18653/v1/D17-1228} {Steering output style
  and topic in neural response generation}.
\newblock In \emph{Proceedings of the 2017 Conference on Empirical Methods in
  Natural Language Processing}, pages 2140--2150. Association for Computational
  Linguistics.

\bibitem[{Wen et~al.(2017)Wen, Vandyke, Mrk{\v{s}}i{\'{c}}, Gasic,
  Rojas~Barahona, Su, Ultes, and Young}]{wen2016network}
Tsung-Hsien Wen, David Vandyke, Nikola Mrk{\v{s}}i{\'{c}}, Milica Gasic,
  Lina~M. Rojas~Barahona, Pei-Hao Su, Stefan Ultes, and Steve Young. 2017.
\newblock \href {http://aclweb.org/anthology/E17-1042} {A network-based
  end-to-end trainable task-oriented dialogue system}.
\newblock In \emph{Proceedings of the 15th Conference of the European Chapter
  of the Association for Computational Linguistics: Volume 1, Long Papers},
  pages 438--449. Association for Computational Linguistics.

\bibitem[{Xing et~al.(2017)Xing, Wu, Wu, Liu, Huang, Zhou, and
  Ma}]{xing2017topic}
Chen Xing, Wei Wu, Yu~Wu, Jie Liu, Yalou Huang, Ming Zhou, and Wei-Ying Ma.
  2017.
\newblock \href
  {https://pdfs.semanticscholar.org/2c0e/1b5db1b6851d95a765a2264bb77f19ee04e1.pdf}
  {Topic aware neural response generation.}
\newblock In \emph{AAAI}, volume~17, pages 3351--3357.

\bibitem[{Zhang et~al.(2018{\natexlab{a}})Zhang, Guo, Fan, Lan, Xu, and
  Cheng}]{zhang2018learning}
Ruqing Zhang, Jiafeng Guo, Yixing Fan, Yanyan Lan, Jun Xu, and Xueqi Cheng.
  2018{\natexlab{a}}.
\newblock \href {http://www.aclweb.org/anthology/P18-1102} {Learning to control
  the specificity in neural response generation}.
\newblock In \emph{Proceedings of the 56th Annual Meeting of the Association
  for Computational Linguistics (Volume 1: Long Papers)}, pages 1108--1117,
  Melbourne, Australia. Association for Computational Linguistics.

\bibitem[{Zhang et~al.(2018{\natexlab{b}})Zhang, Dinan, Urbanek, Szlam, Kiela,
  and Weston}]{Zhang2018PersonalizingToo}
Saizheng Zhang, Emily Dinan, Jack Urbanek, Arthur Szlam, Douwe Kiela, and Jason
  Weston. 2018{\natexlab{b}}.
\newblock \href {http://www.aclweb.org/anthology/P18-1205} {Personalizing
  dialogue agents: I have a dog, do you have pets too?}
\newblock In \emph{Proceedings of the 56th Annual Meeting of the Association
  for Computational Linguistics (Volume 1: Long Papers)}, pages 2204--2213,
  Melbourne, Australia. Association for Computational Linguistics.

\bibitem[{Zhou et~al.(2017)Zhou, Luo, Cao, Lin, Chen, and
  He}]{zhou2017mechanism}
Ganbin Zhou, Ping Luo, Rongyu Cao, Fen Lin, Bo~Chen, and Qing He. 2017.
\newblock \href
  {https://aaai.org/ocs/index.php/AAAI/AAAI17/paper/download/14471/14267}
  {Mechanism-aware neural machine for dialogue response generation}.
\newblock In \emph{AAAI}, pages 3400--3407.

\bibitem[{Zhu et~al.(2015)Zhu, Kiros, Zemel, Salakhutdinov, Urtasun, Torralba,
  and Fidler}]{zhu2015aligning}
Yukun Zhu, Ryan Kiros, Rich Zemel, Ruslan Salakhutdinov, Raquel Urtasun,
  Antonio Torralba, and Sanja Fidler. 2015.
\newblock \href
  {https://www.cv-foundation.org/openaccess/content_iccv_2015/papers/Zhu_Aligning_Books_and_ICCV_2015_paper.pdf}
  {Aligning books and movies: Towards story-like visual explanations by
  watching movies and reading books}.
\newblock In \emph{Proceedings of the IEEE international conference on computer
  vision}, pages 19--27.

\end{thebibliography}
\bibliographystyle{acl_natbib}
  
\onecolumn
\appendix
\section*{\LARGE Supplementary Material}
\vspace{1em}

\section{Screenshots of human evaluation interface}
\label{appendix:human_eval_screenshots}

\begin{figure}[h]
\centering
\includegraphics[width=\textwidth]{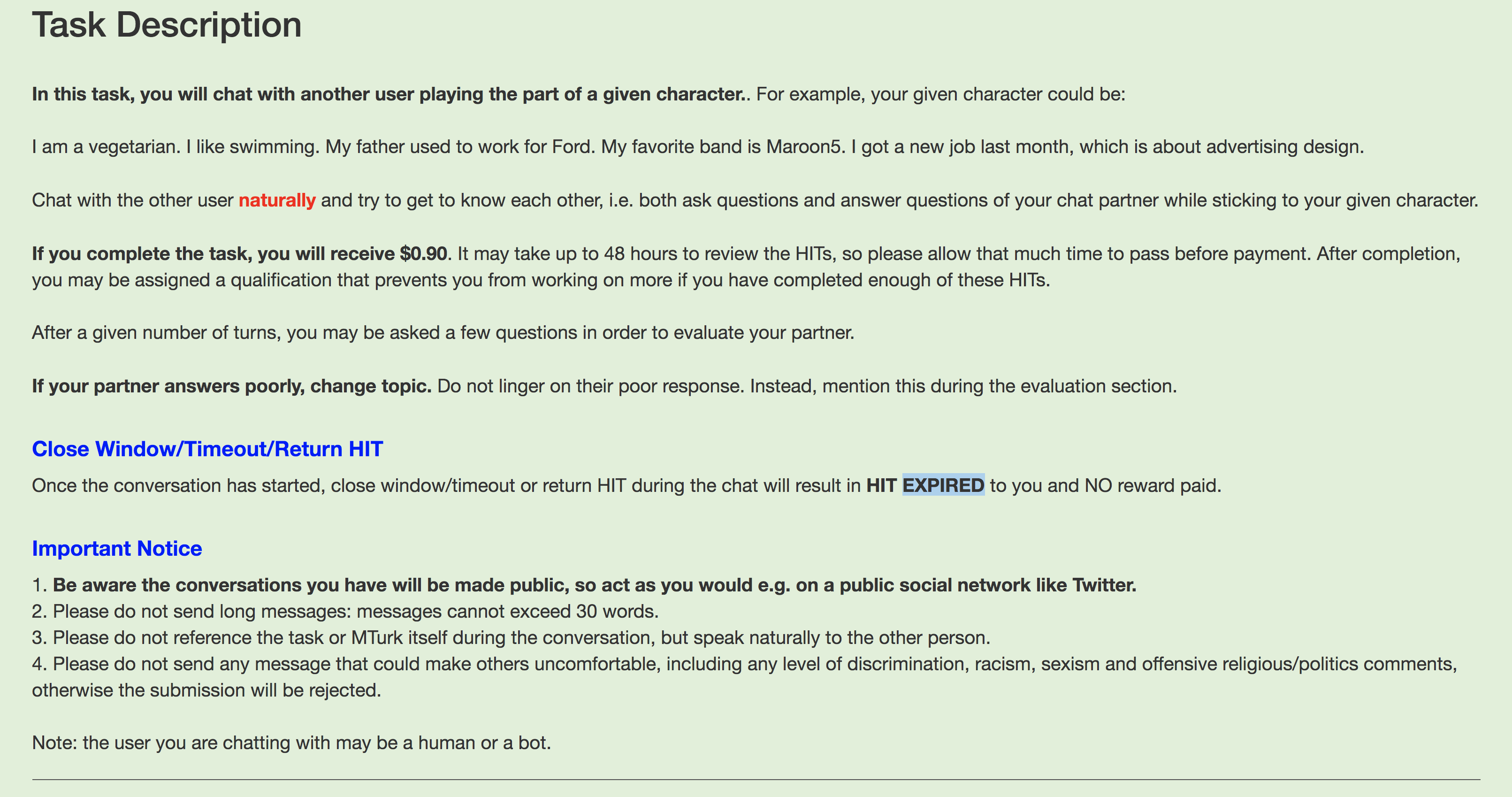}
\caption{Screenshot of the Task Description}
\end{figure}

\begin{figure}[h]
\centering
\includegraphics[width=\textwidth]{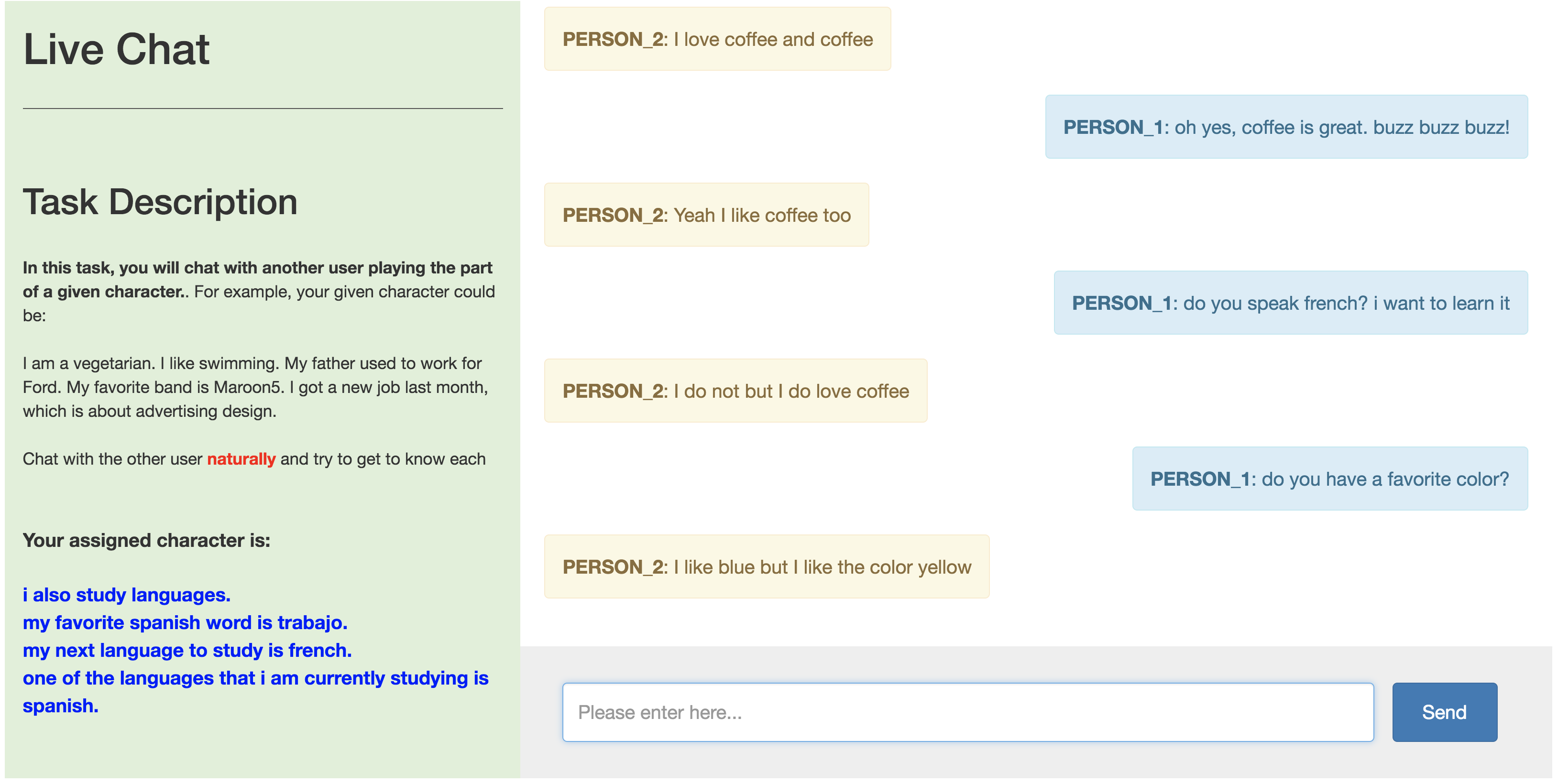}
\caption{Screenshot of the chat UI, talking with the beam search baseline model.}
\end{figure}

\begin{figure}[h]
\centering
\includegraphics[width=\textwidth]{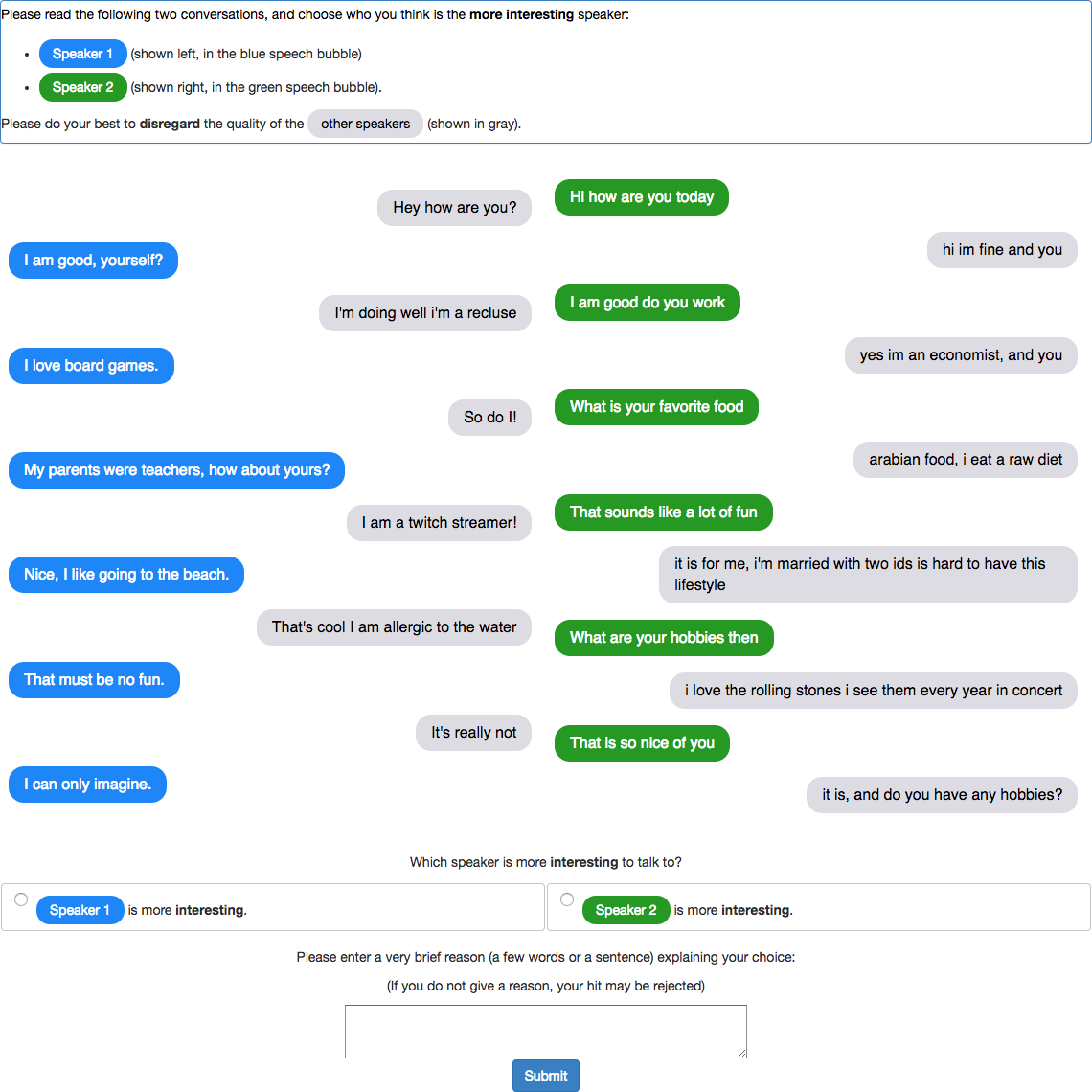}
\caption{Screenshot of the A/B test UI, comparing a human-human conversation (left) and a Repetition-controlled baseline model (right).}
\label{fig:ab_ui}
\end{figure}

\clearpage

\section{Human evaluation questionnaire design}
\centering
\label{appendix:human_eval_questions}
Here are the questions and multiple-choice options used in the human evaluation, in the order presented:\\

\

\fbox{\begin{minipage}{0.95\textwidth}
    {\bf [Engagingness] How much did you enjoy talking to this user?}\\
    $\bullet$ Not at all $\bullet$ A little $\bullet$ Somewhat $\bullet$ A lot \\ \\
    {\bf [Interestingness] How interesting or boring did you find this conversation?}\\
    $\bullet$ Very boring $\bullet$ A little boring $\bullet$ A little interesting $\bullet$ Very interesting \\ \\
    {\bf [Inquisitiveness] How much did the user try to get to know you?}\\
    $\bullet$ Didn't ask about me at all
    $\bullet$ Asked about me some\\
    $\bullet$ Asked about me a good amount
    $\bullet$ Asked about me too much \\ \\
    {\bf [Listening] How much did the user seem to pay attention to what you said?}\\
    $\bullet$ Always ignored what I said $\bullet$ Mostly ignored what I said\\
    $\bullet$ Mostly paid attention to what I said $\bullet$ Always paid attention to what I said \\ \\
    {\bf [Avoiding Repetition] How repetitive was this user?}\\
    $\bullet$ Repeated themselves over and over
    $\bullet$ Sometimes said the same thing twice\\
    $\bullet$ Always said something new \\ \\
    {\bf [Fluency] How naturally did this user speak English?}\\
    $\bullet$ Very unnatural
    $\bullet$ Mostly unnatural
    $\bullet$ Mostly natural
    $\bullet$ Very natural \\ \\
    {\bf [Making sense] How often did this user say something which did NOT make sense?}\\
    $\bullet$ Never made any sense
    $\bullet$ Most responses didn't make sense\\
    $\bullet$ Some responses didn't make sense
    $\bullet$ Everything made perfect sense \\ \\
    {\bf [Humanness] Do you think this user is a bot or a human?}\\
    $\bullet$ Definitely a bot
    $\bullet$ Probably a bot
    $\bullet$ Probably a human
    $\bullet$ Definitely a human \\ \\
    {\bf [Persona retrieval] Which prompt (character) do you think the other user was given for this conversation?} \\
    \textit{Respondent chooses one of two provided personas}
\end{minipage}}

\clearpage
\section{Example conversations from human evaluation}
\label{appendix:conversation_examples}




\begin{figure}[h!]
\centering
\begin{tabular}{cc}
    \frame{\includegraphics[width=0.49\textwidth]{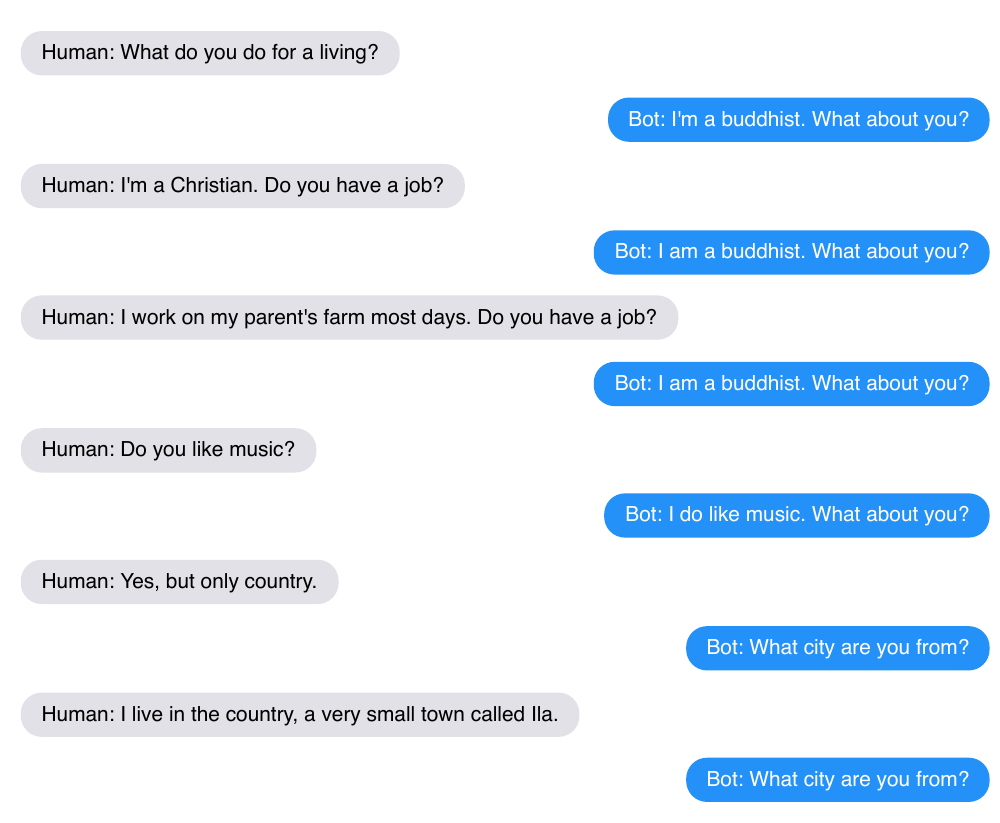}} & \frame{\includegraphics[width=0.49\textwidth]{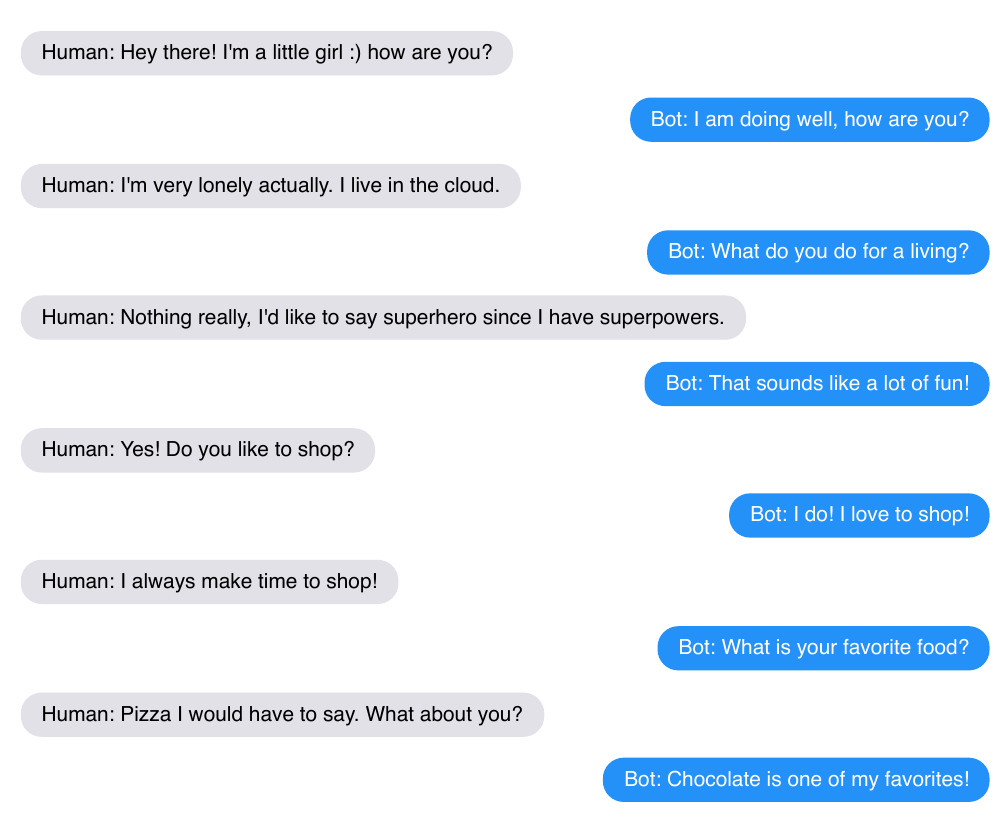}}  \\
    (a) & (b)\\
    &\\
    \frame{\includegraphics[width=0.49\textwidth]{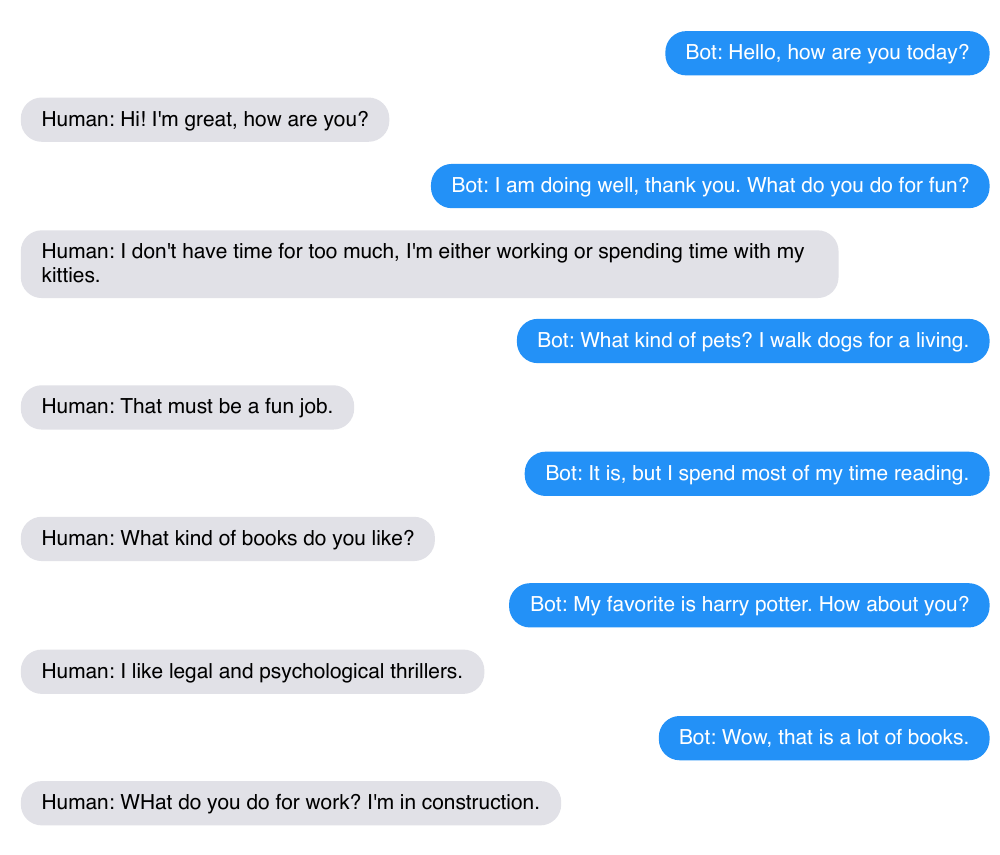}} & \frame{\includegraphics[width=0.49\textwidth]{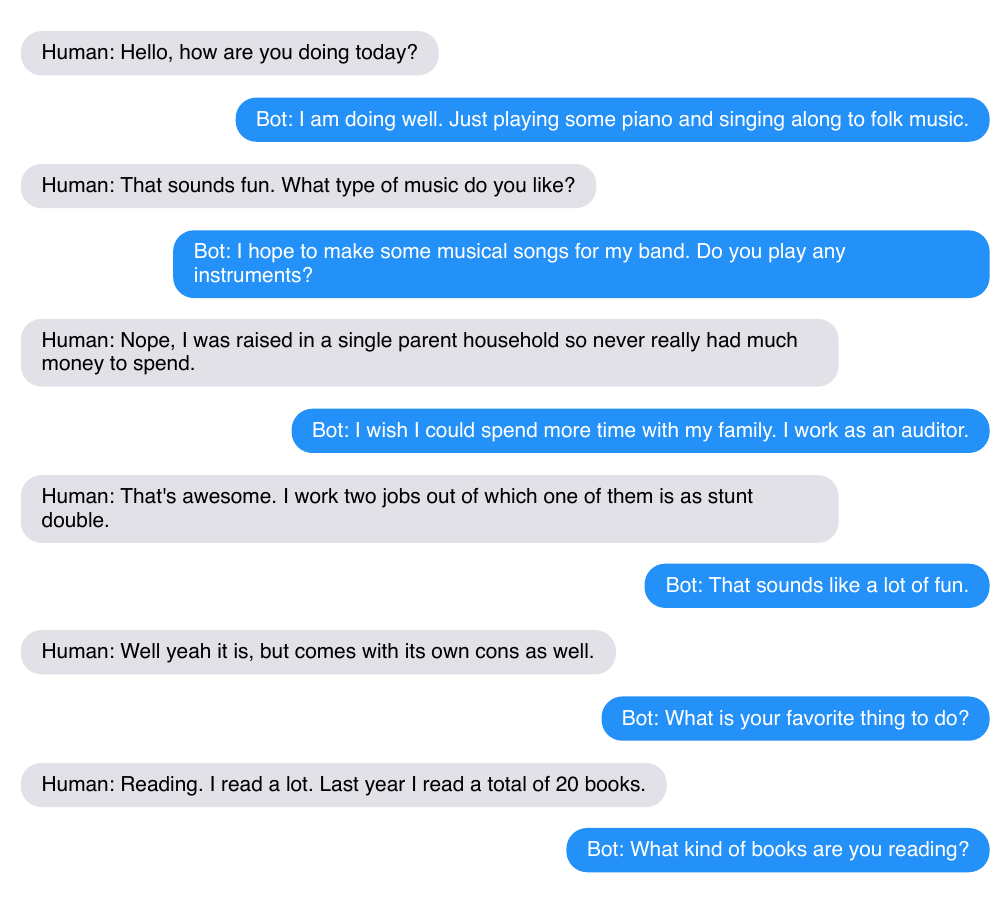}}\\
    (c) & (d)\\
    
\end{tabular}
    \caption{Example conversation with (a) Baseline (b) Repetition-controlled baseline (c) Question-controlled CT ($z=7$), (d) Specificity-controlled WD ($\text{weight}=4$).}
    \label{fig:exbase}
\end{figure}

\clearpage
\section{Repetition-control decoding features}
\label{appendix:rep_feats}

\begin{table}[h!]
\begin{center}
\begin{tabular}{ll}
    \toprule
    \textbf{Feature} & \textbf{Condition} \\
    \midrule
    $\smalltt{extrep\_bigram}(w,y_{<t},x)$ & Adding $w$ to the hypothesis $y_{<t}$ would create a 2-gram \\
    & that appears in a previous utterance by the model \\ \hline
    $\smalltt{extrep\_unigram}(w,y_{<t},x)$ & $w$ is a non-stopword \textbf{and} \\
    & $w$ appears in a previous utterance by the model \\ \hline
    $\smalltt{intrep\_bigram}(w,y_{<t},x)$ & Adding $w$ to the hypothesis $y_{<t}$ would create a 2-gram \\
    & that appears earlier in the hypothesis $y_{<t}$ \\ \hline
    $\smalltt{intrep\_unigram}(w,y_{<t},x)$ & $w$ is a non-stopword \textbf{and} \\ 
    & $w$ appears earlier in the hypothesis $y_{<t}$ \\ \hline
    $\smalltt{partnerrep\_bigram}(w,y_{<t},x)$ & Adding $w$ to the hypothesis $y_{<t}$ would create a 2-gram \\ 
    & that appears in a previous utterance by the partner \\
    \bottomrule
\end{tabular}
\end{center}
\caption{We define five binary features for controlling different types of repetition via weighted decoding (see Section \ref{subsec:wd}). Each feature depends on the word $w$, the partial hypothesis $y_{<t}$, and the context $x$ (which includes the model's own persona and the dialogue history). Each of these features is equal to 1 if and only if the condition on the right is true; otherwise 0.}
\label{tab:rep_feats}
\end{table}

\clearpage
\section{Control settings for all configurations}
\label{appendix:control_settings}

\begin{table}[h!]
\scriptsize
    \centering
   \begin{tabular}{lrrrrrrrr}
   \toprule
& \multicolumn{5}{c}{\textbf{Repetition}} & \textbf{Specificity} & \textbf{Response-rel} & \textbf{Questions} \\ \cmidrule(lr){2-6} \cmidrule(lr){7-7} \cmidrule(lr){8-8} \cmidrule(lr){9-9}
& \multicolumn{2}{c}{\textbf{External}} & \multicolumn{2}{c}{\textbf{Internal}} & \textbf{Partner Rep.} & & & \\ \cmidrule(lr){2-6}
& \textbf{Bigram} & \textbf{Unigram} & \textbf{Bigram} & \textbf{Unigram} & \textbf{Bigram} & \textbf{NIDF} & \textbf{Cos sim} & \textbf{Has `?'} \\
\midrule
\textbf{Baselines} &  &  &  &  &  &  &  & \\
Greedy Search &  &  &  &  &  &  &  & \\
Beam Search (beam size 20) &  &  &  &  &  &  &  & \\
\midrule
\textbf{Repetition control (WD)} &  &  &  &  &  &  &  & \\
Extrep bigram WD -0.5 & wt -0.5 &  &  &  &  &  &  & \\
Extrep bigram WD -1.25 & wt -1.25 &  &  &  &  &  &  & \\
Extrep bigram WD -3.5 & wt -3.5 &  &  &  &  &  &  & \\
Extrep bigram WD -inf & wt -$\infty$ &  &  &  &  &  &  & \\
Repetition-controlled baseline & wt -3.5 &wt -$\infty$ &  &wt -$\infty$ &  &  &  & \\
\midrule
\textbf{Question control (CT)} &  &  &  &  &  &  &  & \\
Question-controlled CT 0 & wt -3.5 &wt -$\infty$ &  &wt -$\infty$ &  &  &  & $z=0$\\
Question-controlled CT 1 & wt -3.5 &wt -$\infty$ &  &wt -$\infty$ &  &  &  & $z=1$\\
Question-controlled CT 4 & wt -3.5 &wt -$\infty$ &  &wt -$\infty$ &  &  &  & $z=4$\\
Question-controlled CT 7 & wt -3.5 &wt -$\infty$ &  &wt -$\infty$ &  &  &  & $z=7$\\
Question-controlled CT 10 & wt -3.5 &wt -$\infty$ &  &wt -$\infty$ &  &  &  & $z=10$\\
Question-controlled CT 10 (boost) & wt 0 * & wt -$\infty$ &  &wt -$\infty$ &  &  &  & $z=10$\\
\midrule
\textbf{Specificity control (CT)} &  &  &  &  &  &  &  & \\
Specificity-controlled CT 0 & wt -3.5 &wt -$\infty$ &  &wt -$\infty$ &  & $z=0$ &  & \\
Specificity-controlled CT 2 & wt -3.5 &wt -$\infty$ &  &wt -$\infty$ &  & $z=2$ &  & \\
Specificity-controlled CT 4 & wt -3.5 &wt -$\infty$ &  &wt -$\infty$ &  & $z=4$ &  & \\
Specificity-controlled CT 7 & wt -3.5 &wt -$\infty$ &  &wt -$\infty$ &  & $z=7$ &  & \\
Specificity-controlled CT 9 & wt -3.5 &wt -$\infty$ &  &wt -$\infty$ &  & $z=9$ &  & \\
\midrule
\textbf{Specificity control (WD)} &  &  &  &  &  &  &  & \\
Specificity-controlled WD -10 & wt -3.5 &wt -$\infty$ &  &wt -$\infty$ &  & wt -10 &  & \\
Specificity-controlled WD -4 & wt -3.5 &wt -$\infty$ &  &wt -$\infty$ &  & wt -4 &  & \\
Specificity-controlled WD 4 & wt -3.5 &wt -$\infty$ &  &wt -$\infty$ &  & wt 4 &  & \\
Specificity-controlled WD 6 & wt -3.5 &wt -$\infty$ &  &wt -$\infty$ &  & wt 6 &  & \\
Specificity-controlled WD 8 & wt -3.5 &wt -$\infty$ &  &wt -$\infty$ &  & wt 8 &  & \\
\midrule
\textbf{Response-related control (WD)} ** &  &  &  &  &  &  &  & \\
Response-related controlled WD -10 & wt -3.5 &wt -$\infty$ &wt -$\infty$ &wt -$\infty$ &wt -$\infty$ &  & wt -10 & \\
Response-related controlled WD 0 & wt -3.5 &wt -$\infty$ &wt -$\infty$ &wt -$\infty$ &wt -$\infty$ &  & wt 0 & \\
Response-related controlled WD 5 & wt -3.5 &wt -$\infty$ &wt -$\infty$ &wt -$\infty$ &wt -$\infty$ &  & wt 5 & \\
Response-related controlled WD 10 & wt -3.5 &wt -$\infty$ &wt -$\infty$ &wt -$\infty$ &wt -$\infty$ &  & wt 10 & \\
Response-related controlled WD 13 & wt -3.5 &wt -$\infty$ &wt -$\infty$ &wt -$\infty$ &wt -$\infty$ &  & wt 13 & \\
\bottomrule
\end{tabular}
\caption{Control settings for all configurations that were human-evaluated. `wt' means the weight used for a weighted decoding feature and `$z=$' means the setting (i.e. bucket) for the control variable in conditional training. \\ \\
* In the setting Question-controlled CT 10 (boost), the feature \smalltt{extrep\_bigram} is \textit{not} used for weighted decoding during beam search, but it \textit{is} used to rerank the candidates after beam search. See Section \ref{subsec:qn} for details. \\ \\
** Note that the Response-related controlled models additionally introduce repetition controls to block internal bigram repetition and partner bigram repetition. This was necessary to prevent the model from parroting the partner's last utterance. In Table~\ref{tab:calibrated}, we find that just adding these extra repetition controls (here called Response-related controlled WD 0, i.e. increased repetition control but no response-relatedness control) outperforms our canonical Repetition-controlled baseline. However, given that we discovered this later, our specificity and question controlled models are built on top of the canonical Repetition-controlled baseline.}
\label{tab:control_settings}
\end{table}

\newpage
\section{Automatic metrics for all configurations}
\label{appendix:auto_metrics}

\begin{table}[h!]
\scriptsize
    \centering
   \begin{tabular}{lrrrrrrrr}
   \toprule
& \multicolumn{5}{c}{\textbf{Repetition}} & \textbf{Specificity} & \textbf{Response-rel} & \textbf{Questions} \\ \cmidrule(lr){2-6} \cmidrule(lr){7-7} \cmidrule(lr){8-8} \cmidrule(lr){9-9}
& \multicolumn{2}{c}{\textbf{External}} & \multicolumn{2}{c}{\textbf{Internal}} & \textbf{Partner Rep.} & & & \\ \cmidrule(lr){2-6}
& \textbf{Bigram} & \textbf{Unigram} & \textbf{Bigram} & \textbf{Unigram} & \textbf{Bigram} & \textbf{NIDF} & \textbf{Cos sim} & \textbf{Has `?'} \\
\midrule
\textbf{Gold data and baselines} &  &  &  &  &  &  &  & \\
Gold Data & 4.65\% & 9.62\% & 0.38\% & 0.97\% & 5.10\% & 0.2119 & 0.1691 & 28.80\% \\ 
Greedy Search & 35.88\% & 36.31\% & 8.08\% & 10.59\% & 12.20\% & 0.1688 & 0.1850 & 6.46\% \\ 
Beam Search (beam size 20) & 46.85\% & 44.15\% & 0.32\% & 0.61\% & 12.90\% & 0.1662 & 0.0957 & 80.87\% \\ 
\midrule
\textbf{Repetition control (WD)} &  &  &  &  &  &  &  & \\
Extrep bigram WD -0.5 & 19.70\% & 16.85\% & 0.26\% & 0.62\% & 11.93\% & 0.1730 & 0.1348 & 73.04\% \\ 
Extrep bigram WD -1.25 & 4.62\% & 4.79\% & 0.40\% & 0.89\% & 10.61\% & 0.1763 & 0.1504 & 61.22\% \\ 
Extrep bigram WD -3.5 & 0.75\% & 4.61\% & 0.47\% & 0.94\% & 9.89\% & 0.1771 & 0.1681 & 48.89\% \\ 
Extrep bigram WD -inf & 0.00\% & 4.74\% & 0.51\% & 1.05\% & 9.56\% & 0.1780 & 0.1711 & 45.98\% \\ 
Repetition-controlled baseline & 0.73\% & 0.00\% & 0.17\% & 0.00\% & 9.55\% & 0.1766 & 0.1676 & 49.98\% \\ 
\midrule
\textbf{Question control (CT)} &  &  &  &  &  &  &  & \\
Question-controlled CT 0 & 0.06\% & 0.00\% & 0.19\% & 0.00\% & 9.20\% & 0.1871 & 0.1753 & 2.01\% \\ 
Question-controlled CT 1 & 0.09\% & 0.00\% & 0.19\% & 0.00\% & 8.66\% & 0.1844 & 0.1722 & 17.33\% \\ 
Question-controlled CT 4 & 0.40\% & 0.00\% & 0.25\% & 0.00\% & 8.53\% & 0.1794 & 0.1713 & 48.88\% \\ 
Question-controlled CT 7 & 0.80\% & 0.00\% & 0.17\% & 0.00\% & 8.48\% & 0.1771 & 0.1724 & 65.65\% \\ 
Question-controlled CT 10 & 1.27\% & 0.00\% & 0.16\% & 0.00\% & 8.48\% & 0.1761 & 0.1728 & 79.67\% \\ 
Question-controlled CT 10 (boost)* & 7.64\% & 0.00\% & 0.03\% & 0.00\% & 10.76\% & 0.1701 & 0.1651 & 99.54\% \\
\midrule
\textbf{Specificity control (CT)} &  &  &  &  &  &  &  & \\
Specificity-controlled CT 0 & 0.60\% & 0.00\% & 0.20\% & 0.00\% & 9.05\% & 0.1478 & 0.1522 & 48.75\% \\ 
Specificity-controlled CT 2 & 0.28\% & 0.00\% & 0.10\% & 0.00\% & 8.37\% & 0.1772 & 0.1833 & 50.57\% \\ 
Specificity-controlled CT 4 & 0.12\% & 0.00\% & 0.08\% & 0.00\% & 7.90\% & 0.1921 & 0.1877 & 29.46\% \\ 
Specificity-controlled CT 7 & 0.02\% & 0.00\% & 0.14\% & 0.00\% & 8.17\% & 0.2156 & 0.1955 & 16.51\% \\ 
Specificity-controlled CT 9 & 0.01\% & 0.00\% & 0.11\% & 0.00\% & 8.01\% & 0.2462 & 0.1990 & 8.50\% \\ 
\midrule
\textbf{Specificity control (WD)} &  &  &  &  &  &  &  & \\
Specificity-controlled WD -10 & 0.14\% & 0.00\% & 10.59\% & 0.00\% & 8.70\% & 0.1107 & 0.0994 & 33.55\% \\ 
Specificity-controlled WD -4 & 0.65\% & 0.00\% & 1.98\% & 0.00\% & 9.95\% & 0.1501 & 0.1398 & 44.92\% \\ 
Specificity-controlled WD 4 & 0.15\% & 0.00\% & 0.19\% & 0.00\% & 7.54\% & 0.2121 & 0.1972 & 45.53\% \\ 
Specificity-controlled WD 6 & 0.07\% & 0.00\% & 0.13\% & 0.00\% & 6.50\% & 0.2546 & 0.2040 & 39.37\% \\ 
Specificity-controlled WD 8 & 0.01\% & 0.00\% & 0.10\% & 0.00\% & 3.40\% & 0.4035 & 0.1436 & 26.68\% \\ 
\midrule
\textbf{Response-related control (WD)} &  &  &  &  &  &  &  & \\
Response-related controlled WD -10 & 0.13\% & 0.00\% & 0.00\% & 0.00\% & 0.00\% & 0.1914 & -0.0921 & 25.71\% \\ 
Response-related controlled WD 0 & 0.24\% & 0.00\% & 0.00\% & 0.00\% & 0.00\% & 0.1785 & 0.1414 & 44.55\% \\ 
Response-related controlled WD 5 & 0.15\% & 0.00\% & 0.00\% & 0.00\% & 0.00\% & 0.1973 & 0.4360 & 39.78\% \\ 
Response-related controlled WD 10 & 0.05\% & 0.00\% & 0.00\% & 0.00\% & 0.00\% & 0.2535 & 0.6653 & 27.56\% \\ 
Response-related controlled WD 13 & 0.02\% & 0.00\% & 0.00\% & 0.00\% & 0.00\% & 0.2999 & 0.7251 & 20.47\% \\ 
\bottomrule
\end{tabular}
\caption{Automatic metrics (computed over validation set) for all model configurations that were human-evaluated.}
    \label{tab:auto_metrics}
\end{table}

\clearpage

\section{Human evaluation results for all configurations}
\label{appendix:human_eval_table}
\begin{table}[h!]
    \centering
\scalebox{0.592}{
\begin{tabular}{lrrrrrrrr|r}
\toprule
Model &    Avoiding Rep. &           Engage &                Fluency &              Humanness &        Inquisitive &        Interesting &              Listening &           Make Sense &      Persona \\
\midrule
\textbf{Human and baselines} & & & & & & & & & \\
Human                              &        2.90 $\pm$ 0.39 &        3.31 $\pm$ 0.90 &        3.66 $\pm$ 0.71 &        3.40 $\pm$ 0.80 &        2.63 $\pm$ 0.63 &        3.23 $\pm$ 0.83 &        3.64 $\pm$ 0.63 &        3.84 $\pm$ 0.52 &        0.92 $\pm$ 0.27 \\
Greedy Search             &        2.16 $\pm$ 0.72 &        2.31 $\pm$ 1.08 &        3.20 $\pm$ 0.81 &        1.78 $\pm$ 0.90 &        2.00 $\pm$ 0.81 &        2.36 $\pm$ 0.98 &        2.78 $\pm$ 0.84 &        3.33 $\pm$ 0.75 &        0.87 $\pm$ 0.34 \\
Beam Search (beam size 20)               &        2.14 $\pm$ 0.72 &        2.35 $\pm$ 1.01 &        3.23 $\pm$ 0.93 &        1.81 $\pm$ 0.87 &        2.50 $\pm$ 0.72 &        2.35 $\pm$ 0.98 &        2.63 $\pm$ 0.85 &        3.40 $\pm$ 0.77 &        0.77 $\pm$ 0.42 \\
\midrule
\textbf{Repetition control (WD)} & & & & & & & & & \\
Extrep bigram WD -0.5                  &        2.66 $\pm$ 0.56 &        2.56 $\pm$ 0.92 &        3.57 $\pm$ 0.64 &        2.19 $\pm$ 0.94 &        2.67 $\pm$ 0.62 &        2.61 $\pm$ 0.87 &        3.08 $\pm$ 0.78 &        3.60 $\pm$ 0.57 &        0.75 $\pm$ 0.43 \\
Extrep bigram WD -1.25                 &        2.84 $\pm$ 0.39 &        2.91 $\pm$ 0.90 &        3.59 $\pm$ 0.64 &        2.32 $\pm$ 0.98 &        2.63 $\pm$ 0.60 &        2.86 $\pm$ 0.89 &        3.21 $\pm$ 0.71 &        3.64 $\pm$ 0.62 &        0.72 $\pm$ 0.45 \\
Extrep bigram WD -3.5                  &        2.90 $\pm$ 0.30 &        2.95 $\pm$ 0.86 &  {\bf 3.73 $\pm$ 0.50} &        2.45 $\pm$ 1.03 &        2.55 $\pm$ 0.61 &        2.88 $\pm$ 0.80 &        3.27 $\pm$ 0.79 &        3.68 $\pm$ 0.49 &        0.80 $\pm$ 0.40 \\
Extrep bigram WD -inf                  &        2.82 $\pm$ 0.43 &        2.96 $\pm$ 0.86 &        3.64 $\pm$ 0.58 &        2.40 $\pm$ 0.96 &        2.65 $\pm$ 0.69 &        2.86 $\pm$ 0.82 &        3.31 $\pm$ 0.69 &        3.66 $\pm$ 0.59 &        0.91 $\pm$ 0.29 \\
Repetition-controlled baseline    &        2.89 $\pm$ 0.39 &        2.89 $\pm$ 0.89 &        3.66 $\pm$ 0.56 &        2.50 $\pm$ 0.99 &        2.70 $\pm$ 0.64 &        2.96 $\pm$ 0.92 &        3.25 $\pm$ 0.71 &        3.68 $\pm$ 0.54 &        0.87 $\pm$ 0.34 \\
\midrule
\textbf{Question control (CT)} & & & & & & & & & \\
Question-controlled CT 0         &        2.95 $\pm$ 0.25 &        2.92 $\pm$ 0.90 &        3.70 $\pm$ 0.54 &        2.49 $\pm$ 0.97 &        2.48 $\pm$ 0.72 &        2.85 $\pm$ 0.93 &        3.29 $\pm$ 0.69 &        3.56 $\pm$ 0.66 &        0.86 $\pm$ 0.35 \\
Question-controlled CT 1         &        2.88 $\pm$ 0.33 &        2.94 $\pm$ 0.93 &        3.59 $\pm$ 0.66 &        2.47 $\pm$ 0.95 &        2.52 $\pm$ 0.69 &        2.85 $\pm$ 0.90 &        3.32 $\pm$ 0.73 &        3.63 $\pm$ 0.55 &        0.85 $\pm$ 0.36 \\
Question-controlled CT 4         &        2.88 $\pm$ 0.38 &        2.88 $\pm$ 0.94 &        3.59 $\pm$ 0.73 &        2.42 $\pm$ 1.07 &        2.55 $\pm$ 0.66 &        2.82 $\pm$ 0.85 &        {\bf 3.37 $\pm$ 0.74} &        3.63 $\pm$ 0.59 &        0.84 $\pm$ 0.37 \\
Question-controlled CT 7         &        2.88 $\pm$ 0.37 &  {\bf 3.07 $\pm$ 0.90} &        3.67 $\pm$ 0.54 &        2.42 $\pm$ 0.98 &        2.75 $\pm$ 0.58 &        2.97 $\pm$ 0.84 &        3.23 $\pm$ 0.76 &        3.53 $\pm$ 0.76 &        0.80 $\pm$ 0.40 \\
Question-controlled CT 10         &        2.74 $\pm$ 0.46 &        2.90 $\pm$ 0.93 &        3.70 $\pm$ 0.50 &        2.43 $\pm$ 1.04 &        2.71 $\pm$ 0.57 &        2.72 $\pm$ 0.88 &        3.12 $\pm$ 0.73 &        3.59 $\pm$ 0.66 &        0.79 $\pm$ 0.41 \\
Question-controlled CT 10 (boost)         &        2.76 $\pm$ 0.49 &        2.84 $\pm$ 0.94 &        3.60 $\pm$ 0.64 &        2.26 $\pm$ 0.97 &  {\bf 2.94 $\pm$ 0.57} &        2.83 $\pm$ 0.94 &        3.18 $\pm$ 0.80 &        3.52 $\pm$ 0.67 &        0.72 $\pm$ 0.45 \\
\midrule
\textbf{Specificity control (CT)} & & & & & & & & & \\
Specificity-controlled CT 0      &        2.83 $\pm$ 0.40 &        2.96 $\pm$ 0.93 &        3.62 $\pm$ 0.58 &        2.42 $\pm$ 0.99 &        2.60 $\pm$ 0.56 &        2.86 $\pm$ 0.89 &        3.29 $\pm$ 0.70 &        3.66 $\pm$ 0.60 &        0.72 $\pm$ 0.45 \\
Specificity-controlled CT 2      &        2.90 $\pm$ 0.36 &        2.78 $\pm$ 1.00 &        3.60 $\pm$ 0.64 &        2.37 $\pm$ 0.93 &        2.66 $\pm$ 0.66 &        2.80 $\pm$ 0.96 &        3.14 $\pm$ 0.77 &        3.50 $\pm$ 0.63 &        0.81 $\pm$ 0.39 \\
Specificity-controlled CT 4      &        2.92 $\pm$ 0.27 &        2.81 $\pm$ 0.88 &        3.65 $\pm$ 0.59 &        2.34 $\pm$ 1.02 &        2.57 $\pm$ 0.62 &        2.80 $\pm$ 0.78 &        3.25 $\pm$ 0.78 &        3.50 $\pm$ 0.66 &        0.86 $\pm$ 0.35 \\
Specificity-controlled CT 7      &        2.89 $\pm$ 0.32 &        3.00 $\pm$ 0.94 &        3.64 $\pm$ 0.67 &        2.53 $\pm$ 1.03 &        2.56 $\pm$ 0.66 &        2.90 $\pm$ 0.90 &        3.34 $\pm$ 0.70 &        3.59 $\pm$ 0.60 &        0.82 $\pm$ 0.39 \\
Specificity-controlled CT 9      &        2.90 $\pm$ 0.35 &        2.83 $\pm$ 0.87 &        3.61 $\pm$ 0.62 &        2.40 $\pm$ 0.97 &        2.31 $\pm$ 0.74 &        2.84 $\pm$ 0.83 &        3.07 $\pm$ 0.81 &        3.58 $\pm$ 0.56 &        0.88 $\pm$ 0.32 \\
\midrule
\textbf{Specificity control (WD)} & & & & & & & & & \\
Specificity-controlled WD -10      &        2.85 $\pm$ 0.43 &        2.43 $\pm$ 0.99 &        3.34 $\pm$ 0.83 &        2.15 $\pm$ 0.91 &        2.31 $\pm$ 0.69 &        2.38 $\pm$ 0.94 &        3.03 $\pm$ 0.75 &        3.33 $\pm$ 0.70 &        0.71 $\pm$ 0.45 \\
Specificity-controlled WD -4      &        2.90 $\pm$ 0.30 &        2.78 $\pm$ 0.95 &        3.55 $\pm$ 0.63 &        2.41 $\pm$ 0.92 &        2.52 $\pm$ 0.66 &        2.64 $\pm$ 0.93 &        3.28 $\pm$ 0.73 &        3.56 $\pm$ 0.62 &        0.82 $\pm$ 0.38 \\
Specificity-controlled WD 4      &        2.95 $\pm$ 0.21 &        2.99 $\pm$ 0.86 &        3.65 $\pm$ 0.55 &        2.49 $\pm$ 0.90 &        2.65 $\pm$ 0.55 &        3.00 $\pm$ 0.78 &        {\bf 3.37 $\pm$ 0.59} &        3.63 $\pm$ 0.50 &  {\bf 0.93 $\pm$ 0.25} \\
Specificity-controlled WD 6      &        2.93 $\pm$ 0.26 &        2.96 $\pm$ 0.90 &        3.52 $\pm$ 0.76 &        2.41 $\pm$ 1.04 &        2.58 $\pm$ 0.66 &  {\bf 3.06 $\pm$ 0.80} &        3.24 $\pm$ 0.76 &        3.50 $\pm$ 0.66 &        {\bf 0.93 $\pm$ 0.26} \\
Specificity-controlled WD 8      &        2.78 $\pm$ 0.52 &        2.40 $\pm$ 1.23 &        2.67 $\pm$ 1.25 &        1.86 $\pm$ 0.97 &        2.03 $\pm$ 0.87 &        2.55 $\pm$ 1.14 &        2.61 $\pm$ 1.05 &        2.91 $\pm$ 0.91 &        0.92 $\pm$ 0.28 \\
\midrule
\textbf{Response-related control (WD)} & & & & & & & & & \\
Response-related controlled WD -10 &        2.86 $\pm$ 0.44 &        2.48 $\pm$ 0.98 &        3.42 $\pm$ 0.74 &        2.02 $\pm$ 0.93 &        2.38 $\pm$ 0.75 &        2.53 $\pm$ 0.94 &        2.84 $\pm$ 0.80 &        3.14 $\pm$ 0.75 &        0.91 $\pm$ 0.29 \\
Response-related controlled WD 0 &  {\bf 2.96 $\pm$ 0.23} &        3.01 $\pm$ 0.90 &        3.72 $\pm$ 0.54 &  {\bf 2.73 $\pm$ 1.00} &        2.56 $\pm$ 0.67 &        2.92 $\pm$ 0.84 &  {\bf 3.37 $\pm$ 0.72} &  {\bf 3.73 $\pm$ 0.52} &        0.82 $\pm$ 0.38 \\
Response-related controlled WD 5 &        2.90 $\pm$ 0.33 &        2.88 $\pm$ 0.90 &        3.51 $\pm$ 0.63 &        2.41 $\pm$ 1.01 &        2.53 $\pm$ 0.65 &        2.85 $\pm$ 0.90 &        3.27 $\pm$ 0.73 &        3.49 $\pm$ 0.63 &        0.82 $\pm$ 0.39 \\
Response-related controlled WD 10 &        2.78 $\pm$ 0.43 &        2.39 $\pm$ 1.04 &        3.06 $\pm$ 0.90 &        1.97 $\pm$ 0.99 &        2.22 $\pm$ 0.67 &        2.57 $\pm$ 1.01 &        3.03 $\pm$ 0.76 &        3.16 $\pm$ 0.63 &        0.75 $\pm$ 0.43 \\
Response-related controlled WD 13 &        2.71 $\pm$ 0.57 &        2.10 $\pm$ 1.13 &        2.54 $\pm$ 1.12 &        1.81 $\pm$ 1.07 &        2.14 $\pm$ 0.84 &        2.33 $\pm$ 1.06 &        2.69 $\pm$ 0.83 &        2.70 $\pm$ 0.88 &        0.62 $\pm$ 0.49 \\
\bottomrule
\end{tabular}
}
\caption{Raw scores (mean $\pm$ std.) for all models and human evaluation metrics. \\ \\
The first eight columns are Likert metrics on a 1-4 scale (except Avoiding Repetition, which is a 1-3 scale),
where higher is better (except Inquisitiveness, which has an optimal score of 3). The last column, Persona Retrieval, is on a scale from 0 to 1 where higher is better. \\ \\
The maximum of each column (excluding Human row) is in bold.}
\label{tab:raw}
\end{table}
\begin{table}[h!]
\centering
\scalebox{0.645}{
\begin{tabular}{lrrrrrrrr}
\toprule
Model &    Avoiding Rep. &           Engage &                Fluency &              Humanness &        Inquisitive &        Interesting &              Listening &           Make Sense \\
\midrule
\textbf{Human and baselines} & & & & & & & & \\
* Human                              &        2.79 $\pm$ 0.12 &        3.04 $\pm$ 0.11 &        3.36 $\pm$ 0.12 &        3.35 $\pm$ 0.11 &        2.44 $\pm$ 0.12 &        2.92 $\pm$ 0.11 &        3.32 $\pm$ 0.13 &        3.68 $\pm$ 0.11 \\
* Greedy Search             &        2.08 $\pm$ 0.10 &        2.24 $\pm$ 0.11 &        3.03 $\pm$ 0.10 &        1.75 $\pm$ 0.12 &        1.95 $\pm$ 0.10 &        2.29 $\pm$ 0.13 &        2.62 $\pm$ 0.10 &        3.23 $\pm$ 0.10 \\
* Beam Search (beam size 20)               &        2.08 $\pm$ 0.11 &        2.29 $\pm$ 0.11 &        3.09 $\pm$ 0.13 &        1.71 $\pm$ 0.13 &        2.42 $\pm$ 0.11 &        2.29 $\pm$ 0.14 &        2.47 $\pm$ 0.12 &        3.35 $\pm$ 0.13 \\
\midrule
\textbf{Repetition control (WD)} & & & & & & & & \\
Extrep bigram WD -0.5                  &        2.62 $\pm$ 0.10 &        2.54 $\pm$ 0.12 &        3.35 $\pm$ 0.12 &        2.13 $\pm$ 0.11 &        2.63 $\pm$ 0.11 &        2.56 $\pm$ 0.11 &        2.93 $\pm$ 0.11 &        3.48 $\pm$ 0.11 \\
Extrep bigram WD -1.25                 &        2.78 $\pm$ 0.09 &        2.82 $\pm$ 0.13 &        3.40 $\pm$ 0.12 &        2.27 $\pm$ 0.12 &        2.54 $\pm$ 0.09 &        2.76 $\pm$ 0.10 &        3.05 $\pm$ 0.11 &        3.53 $\pm$ 0.14 \\
Extrep bigram WD -3.5                  &        2.83 $\pm$ 0.11 &        2.93 $\pm$ 0.10 &  {\bf 3.56 $\pm$ 0.10} &        2.43 $\pm$ 0.11 &        2.47 $\pm$ 0.11 &        2.83 $\pm$ 0.10 &        3.14 $\pm$ 0.10 &        3.62 $\pm$ 0.12 \\
Extrep bigram WD -inf                  &        2.74 $\pm$ 0.11 &        2.87 $\pm$ 0.14 &        3.49 $\pm$ 0.12 &        2.32 $\pm$ 0.13 &        2.56 $\pm$ 0.11 &        2.75 $\pm$ 0.12 &        3.13 $\pm$ 0.12 &        3.59 $\pm$ 0.12 \\
* Repetition-controlled baseline     &        2.86 $\pm$ 0.12 &        2.82 $\pm$ 0.12 &        3.53 $\pm$ 0.10 &        2.40 $\pm$ 0.11 &        2.62 $\pm$ 0.13 &        2.84 $\pm$ 0.12 &        3.10 $\pm$ 0.11 &        3.58 $\pm$ 0.14 \\
\midrule
\textbf{Question control (CT)} & & & & & & & & \\
Question-controlled CT 0         &        {\bf 2.87 $\pm$ 0.12} &        2.84 $\pm$ 0.13 &        3.51 $\pm$ 0.10 &        2.46 $\pm$ 0.11 &        2.36 $\pm$ 0.09 &        2.76 $\pm$ 0.09 &        3.10 $\pm$ 0.10 &        3.49 $\pm$ 0.12 \\
Question-controlled CT 1         &        2.82 $\pm$ 0.11 &        2.88 $\pm$ 0.11 &        3.42 $\pm$ 0.10 &        2.46 $\pm$ 0.12 &        2.47 $\pm$ 0.11 &        2.79 $\pm$ 0.13 &        3.14 $\pm$ 0.11 &        3.55 $\pm$ 0.10 \\
Question-controlled CT 4         &        2.78 $\pm$ 0.12 &        2.88 $\pm$ 0.10 &        3.47 $\pm$ 0.11 &        2.40 $\pm$ 0.09 &        2.53 $\pm$ 0.13 &        2.83 $\pm$ 0.13 &  {\bf 3.24 $\pm$ 0.11} &        3.59 $\pm$ 0.10 \\
* Question-controlled CT 7         &        2.81 $\pm$ 0.10 &  {\bf 2.99 $\pm$ 0.11} &        3.54 $\pm$ 0.09 &        2.35 $\pm$ 0.11 &        2.66 $\pm$ 0.12 &        2.92 $\pm$ 0.12 &        3.11 $\pm$ 0.10 &        3.47 $\pm$ 0.10 \\
Question-controlled CT 10         &        2.67 $\pm$ 0.13 &        2.87 $\pm$ 0.11 &        3.52 $\pm$ 0.12 &        2.35 $\pm$ 0.12 &        2.63 $\pm$ 0.12 &        2.66 $\pm$ 0.10 &        2.94 $\pm$ 0.11 &        3.53 $\pm$ 0.12 \\
Question-controlled CT 10 (boost)          &        2.68 $\pm$ 0.12 &        2.74 $\pm$ 0.09 &        3.42 $\pm$ 0.12 &        2.19 $\pm$ 0.13 &  {\bf 2.79 $\pm$ 0.11} &        2.74 $\pm$ 0.11 &        3.00 $\pm$ 0.12 &        3.45 $\pm$ 0.13 \\
\midrule
\textbf{Specificity control (CT)} & & & & & & & & \\
Specificity-controlled CT 0      &        2.79 $\pm$ 0.10 &        2.93 $\pm$ 0.09 &        3.44 $\pm$ 0.12 &        2.38 $\pm$ 0.11 &        2.56 $\pm$ 0.12 &        2.84 $\pm$ 0.12 &        3.12 $\pm$ 0.13 &        3.61 $\pm$ 0.11 \\
Specificity-controlled CT 2      &        2.78 $\pm$ 0.12 &        2.74 $\pm$ 0.11 &        3.39 $\pm$ 0.13 &        2.31 $\pm$ 0.13 &        2.56 $\pm$ 0.13 &        2.74 $\pm$ 0.12 &        2.99 $\pm$ 0.11 &        3.47 $\pm$ 0.10 \\
Specificity-controlled CT 4      &        2.82 $\pm$ 0.10 &        2.80 $\pm$ 0.13 &        3.44 $\pm$ 0.14 &        2.32 $\pm$ 0.13 &        2.51 $\pm$ 0.12 &        2.78 $\pm$ 0.15 &        3.09 $\pm$ 0.13 &        3.46 $\pm$ 0.13 \\
Specificity-controlled CT 7      &        2.81 $\pm$ 0.12 &        2.91 $\pm$ 0.13 &        3.43 $\pm$ 0.11 &        2.45 $\pm$ 0.10 &        2.49 $\pm$ 0.11 &        2.81 $\pm$ 0.12 &        3.15 $\pm$ 0.12 &        3.55 $\pm$ 0.11 \\
Specificity-controlled CT 9      &        2.80 $\pm$ 0.13 &        2.78 $\pm$ 0.10 &        3.41 $\pm$ 0.12 &        2.35 $\pm$ 0.13 &        2.28 $\pm$ 0.11 &        2.79 $\pm$ 0.11 &        2.91 $\pm$ 0.11 &        3.51 $\pm$ 0.12 \\
\midrule
\textbf{Specificity control (WD)} & & & & & & & & \\
Specificity-controlled WD -10      &        2.76 $\pm$ 0.11 &        2.41 $\pm$ 0.12 &        3.19 $\pm$ 0.12 &        2.15 $\pm$ 0.11 &        2.28 $\pm$ 0.13 &        2.35 $\pm$ 0.12 &        2.89 $\pm$ 0.11 &        3.28 $\pm$ 0.12 \\
Specificity-controlled WD -4      &        2.83 $\pm$ 0.10 &        2.76 $\pm$ 0.12 &        3.37 $\pm$ 0.10 &        2.36 $\pm$ 0.11 &        2.46 $\pm$ 0.11 &        2.62 $\pm$ 0.12 &        3.14 $\pm$ 0.09 &        3.52 $\pm$ 0.11 \\
* Specificity-controlled WD 4      &        2.84 $\pm$ 0.10 &        2.96 $\pm$ 0.12 &        3.45 $\pm$ 0.13 &        2.44 $\pm$ 0.12 &        2.56 $\pm$ 0.09 &  {\bf 2.94 $\pm$ 0.11} &        3.20 $\pm$ 0.10 &        3.54 $\pm$ 0.11 \\
Specificity-controlled WD 6      &        2.81 $\pm$ 0.09 &        2.91 $\pm$ 0.10 &        3.34 $\pm$ 0.09 &        2.31 $\pm$ 0.11 &        2.53 $\pm$ 0.12 &        2.93 $\pm$ 0.12 &        3.09 $\pm$ 0.10 &        3.41 $\pm$ 0.12 \\
Specificity-controlled WD 8      &        2.70 $\pm$ 0.11 &        2.39 $\pm$ 0.12 &        2.54 $\pm$ 0.12 &        1.80 $\pm$ 0.13 &        2.00 $\pm$ 0.10 &        2.49 $\pm$ 0.12 &        2.47 $\pm$ 0.10 &        2.87 $\pm$ 0.11 \\
\midrule
\textbf{Response-related control (WD)} & & & & & & & & \\
Response-related controlled WD -10 &        2.77 $\pm$ 0.12 &        2.45 $\pm$ 0.12 &        3.26 $\pm$ 0.11 &        1.96 $\pm$ 0.10 &        2.31 $\pm$ 0.12 &        2.47 $\pm$ 0.12 &        2.73 $\pm$ 0.11 &        3.12 $\pm$ 0.12 \\
Response-related controlled WD 0 &  {\bf 2.87 $\pm$ 0.12} &        2.97 $\pm$ 0.11 &        3.55 $\pm$ 0.09 &  {\bf 2.62 $\pm$ 0.11} &        2.48 $\pm$ 0.10 &        2.88 $\pm$ 0.12 &        3.21 $\pm$ 0.09 &  {\bf 3.70 $\pm$ 0.10} \\
Response-related controlled WD 5 &        2.79 $\pm$ 0.10 &        2.83 $\pm$ 0.09 &        3.35 $\pm$ 0.12 &        2.40 $\pm$ 0.12 &        2.51 $\pm$ 0.13 &        2.80 $\pm$ 0.13 &        3.13 $\pm$ 0.12 &        3.41 $\pm$ 0.12 \\
Response-related controlled WD 10 &        2.74 $\pm$ 0.11 &        2.42 $\pm$ 0.12 &        2.93 $\pm$ 0.11 &        1.95 $\pm$ 0.12 &        2.20 $\pm$ 0.12 &        2.56 $\pm$ 0.12 &        2.90 $\pm$ 0.12 &        3.12 $\pm$ 0.10 \\
Response-related controlled WD 13 &        2.63 $\pm$ 0.12 &        2.06 $\pm$ 0.11 &        2.40 $\pm$ 0.09 &        1.74 $\pm$ 0.11 &        2.07 $\pm$ 0.11 &        2.25 $\pm$ 0.12 &        2.49 $\pm$ 0.14 &        2.63 $\pm$ 0.10 \\
\bottomrule
\end{tabular}
}
\caption{Calibrated scores (mean $\pm$ std.) for all models and human evaluation metrics. \\ \\
The first eight columns are Likert metrics on a 1-4 scale (except Avoiding Repetition, which is a 1-3 scale),
where higher is better (except Inquisitiveness, which has an optimal score of 3). The last column, Persona Retrieval, is on a scale from 0 to 1 where higher is better. \\ \\
The maximum of each column (excluding Human row) is in bold. \\ \\
Rows marked with * are the six models included in Figure \ref{fig:engage_results} (left) and Figure \ref{fig:birthday}.}
\label{tab:calibrated}
\end{table}

\clearpage

\section{Plots of human evaluation results for all configurations}
\label{appendix:human_eval_plots}

\begin{figure}[h!]
\centering
\includegraphics[width=0.93\textwidth]{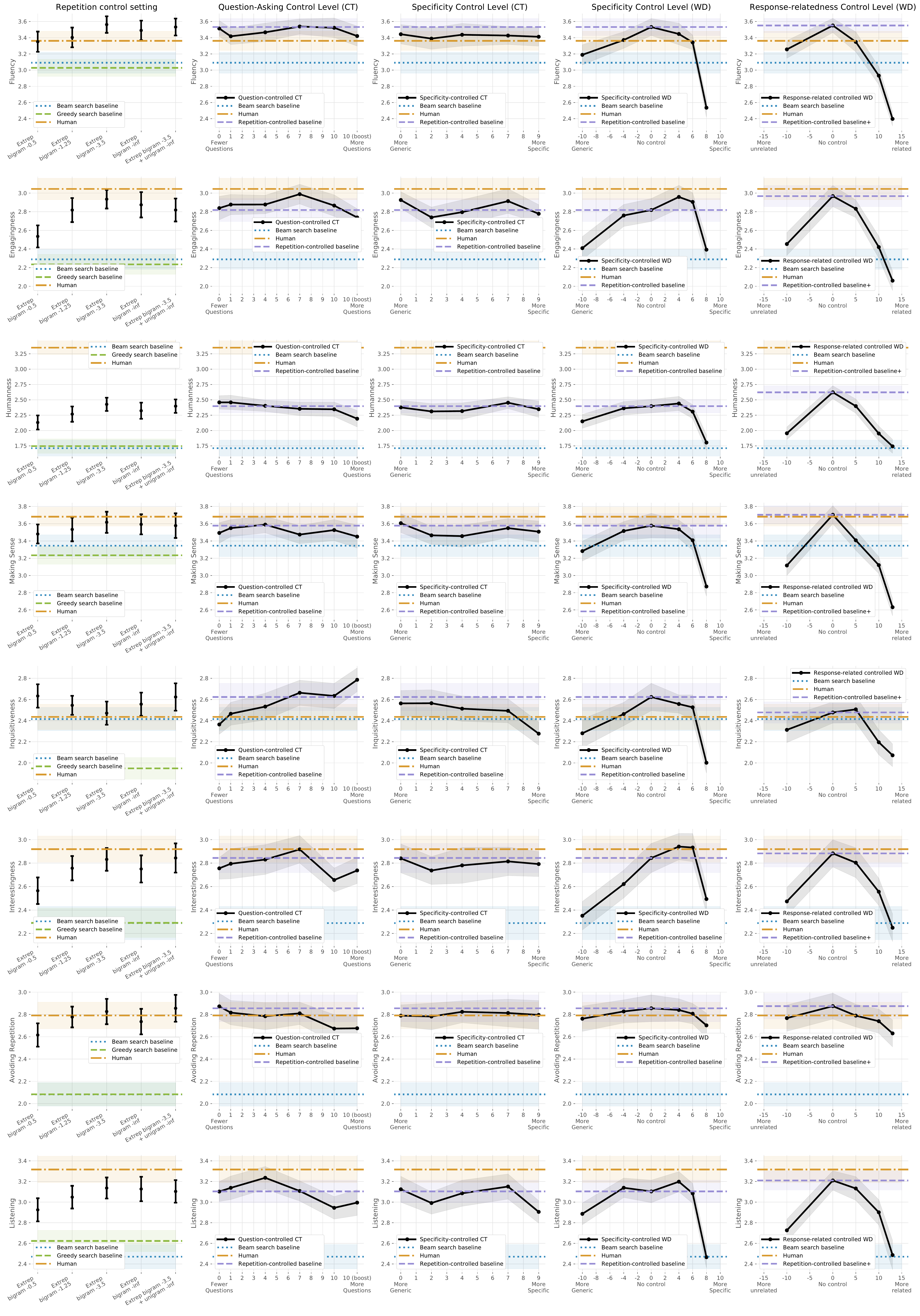}
\caption{Calibrated human evaluation scores for all models. This is the same data as in Table \ref{tab:calibrated}. \\ \\ 
Note: `Repetition-controlled baseline+' in the rightmost column is `Response-related controlled WD 0' in Table \ref{tab:calibrated}. See Table \ref{tab:control_settings} for explanation.}
\label{fig:allplots}
\end{figure}

\end{document}